\documentclass[accepted]{article}

\usepackage{microtype}
\usepackage{graphicx}
\usepackage{subfigure}
\usepackage{booktabs} 

\usepackage{hyperref}
\usepackage{multirow}
\usepackage{diagbox}

\usepackage{icml2024}

\usepackage{amsmath}
\usepackage{amssymb}
\usepackage{mathtools}
\usepackage{amsthm}

\usepackage[capitalize,noabbrev]{cleveref}

\theoremstyle{plain}
\newtheorem{theorem}{Theorem}[section]

\theoremstyle{definition}
\newtheorem{definition}[theorem]{Definition}

\theoremstyle{remark}

\usepackage[textsize=tiny]{todonotes}

\icmltitlerunning{Balancing Similarity and Complementarity for Federated Learning}

\begin{document}

\twocolumn[
\icmltitle{Balancing Similarity and Complementarity for Federated Learning}



\icmlsetsymbol{equal}{*}
\icmlsetsymbol{co}{\dag}

\begin{icmlauthorlist}
\icmlauthor{Kunda Yan}{equal,aa}
\icmlauthor{Sen Cui}{equal,aa}
\icmlauthor{Abudukelimu Wuerkaixi}{aa}
\icmlauthor{Jingfeng Zhang}{bb,cc}
\icmlauthor{Bo Han}{dd,cc}
\icmlauthor{Gang Niu}{cc}
\icmlauthor{Masashi Sugiyama}{co,cc,ee}
\icmlauthor{Changshui Zhang}{co,aa}
\end{icmlauthorlist}

\icmlaffiliation{aa}{ Institute for Artificial Intelligence, Tsinghua University (THUAI), 
Beijing National Research Center for Information Science and Technology (BNRist), 
Department of Automation, Tsinghua University, Beijing, P.R.China}
\icmlaffiliation{bb}{The University of Auckland}
\icmlaffiliation{cc}{RIKEN}
\icmlaffiliation{dd}{Hong Kong Baptist University}
\icmlaffiliation{ee}{The University of Tokyo}

\icmlcorrespondingauthor{Masashi Sugiyama}{sugi@k.u-tokyo.ac.jp}
\icmlcorrespondingauthor{Changshui Zhang}{zcs@mail.tsinghua.edu.cn}

\icmlkeywords{Machine Learning, ICML}

\vskip 0.3in
]



\printAffiliationsAndNotice{\icmlEqualContribution} 

\begin{abstract}

In mobile and IoT systems, Federated Learning (FL) is increasingly important for effectively using data while maintaining user privacy. One key challenge in FL is managing statistical heterogeneity, such as non-i.i.d. data, arising from numerous clients and diverse data sources. This requires strategic cooperation, often with clients having similar characteristics. However, we are interested in a fundamental question: does achieving optimal cooperation necessarily entail cooperating with the most similar clients? Typically, significant model performance improvements are often realized not by partnering with the most similar models, but through leveraging complementary data.  Our theoretical and empirical analyses suggest that optimal cooperation is achieved by enhancing complementarity in feature distribution while restricting the disparity in the correlation between features and targets. Accordingly, we introduce a novel framework, \texttt{FedSaC}, which balances similarity and complementarity in FL cooperation. Our framework aims to approximate an optimal cooperation network for each client by optimizing a weighted sum of model similarity and feature complementarity. The strength of \texttt{FedSaC} lies in its adaptability to various levels of data heterogeneity and multimodal scenarios. Our comprehensive unimodal and multimodal experiments demonstrate that \texttt{FedSaC} markedly surpasses other state-of-the-art FL methods.
\end{abstract}

\section{Introduction}
\label{Introduction}

Federated Learning (FL)~\cite{DBLP:conf/aistats/McMahanMRHA17}, emerging as a pivotal paradigm in machine learning, is increasingly acclaimed for  facilitating collaborative training across diverse clients while ensuring data confidentiality. However, FL still encounters significant challenges, chiefly statistical heterogeneity - the occurrence of non-i.i.d. data across diverse local clients, as explored in prior research.~\cite{DBLP:conf/kdd/CuiLPCZW22, DBLP:conf/cvpr/QuZLXW00R22, DBLP:conf/icml/KarimireddyKMRS20, DBLP:conf/cvpr/LiSAS23}.
In real-world scenarios with data from heterogeneous user bases, models often face performance decline due to local data distribution variances~\cite{DBLP:journals/ftml/KairouzMABBBBCC21, DBLP:conf/icde/LiDCH22, DBLP:conf/cvpr/HuangY022}.

In the context of multimodal learning, statistical heterogeneity is notably pronounced~\cite{DBLP:conf/kdd/ChenZ20, DBLP:conf/mobicom/ZhengLCW023}. Variations in dimensionality, quality, and reliability among diverse data sources exacerbate heterogeneity within each client's modalities and magnify distribution discrepancies between clients. Such significant heterogeneity complicates achieving consistent and efficient learning in the FL framework~\cite{DBLP:conf/iclr/YuLWXL23, DBLP:journals/ijautcomp/LinGGZZL23}.

In response to this challenge, a promising direction involves the identification of optimal collaborators  predicated on model similarity metrics~\cite{DBLP:conf/icml/BaekJJYH23, DBLP:journals/tnn/SattlerMS21, DBLP:conf/icml/YeNWCW23}. For example,  IFCA~\cite{DBLP:conf/icml/BaekJJYH23} clusters cooperative clients based on the similarity of their model parameters, whereas CFL~\cite{DBLP:journals/tnn/SattlerMS21} employs gradient similarity for the same purpose. pFedGraph~\cite{DBLP:conf/icml/YeNWCW23} constructs a cooperation graph guided by an intuitive notion that clients with greater similarity should collaborate more intensively. 
These methods collectively emphasize the importance of model similarity in strategic collaboration.

\begin{figure*}[htbp]
\setlength{\abovecaptionskip}{-.1cm}
\setlength{\belowcaptionskip}{-.3cm}
\centering{
\includegraphics[width=2\columnwidth]{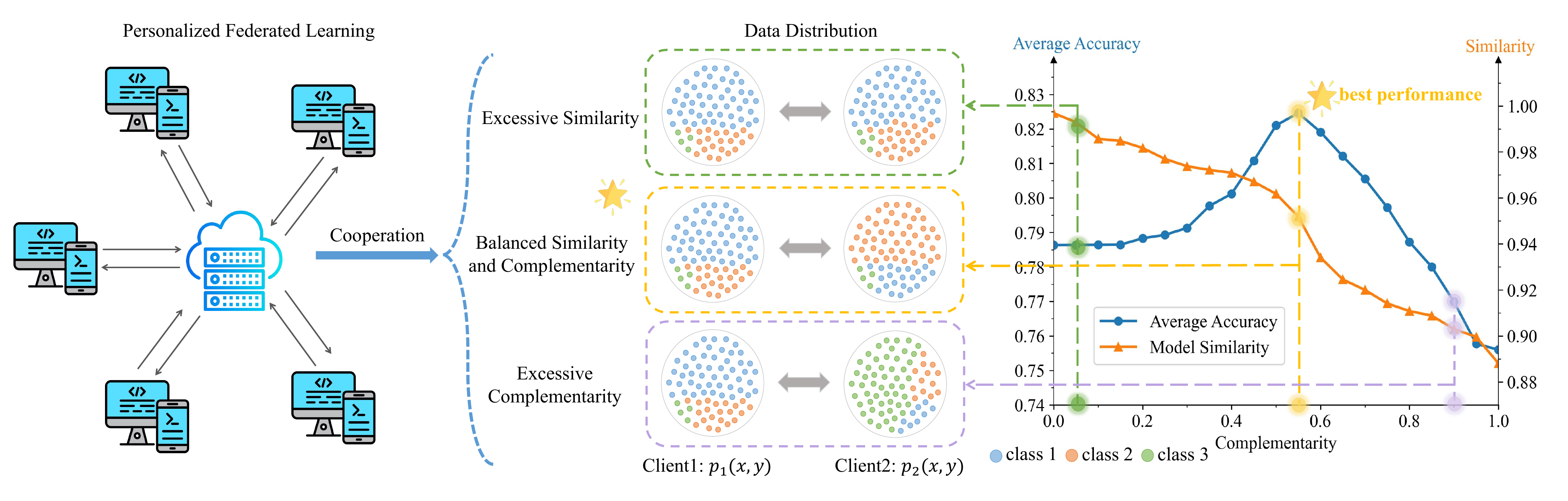}}
\vspace{-.2cm}
\caption{Illustration of the role of data complementarity on personalized federated learning cooperation. The figure presents the experimental results how increasing data complementarity between two clients influences average accuracy post-cooperation and model similarity. Three scenarios are presented, showcasing distinct levels of complementarity for local data distributions. The findings underscore the benefits of complementarity, revealing that a balance of similarity and complementarity enhances cooperative benefits in a federated learning framework. 
}
\label{fig:intro}
\vspace{-.3cm}
\end{figure*}

However, we raise a fundamental question: \emph{does achieving optimal cooperation necessarily entail cooperating with the most similar clients?} Theoretically, similarity oriented from collaboration is conservative, such that it could potentially result in unproductive cooperation. For example, under the assumption of two completely identical clients or models, cooperation between them would yield no information gain for either party, despite their maximal similarity.
Interestingly, the fundamental precondition for model enhancement through collaboration is complementarity, not similarity.

Inspired by this intuition, we design experiments to explore  the underlying mechanism. As an illustration in FL, we exemplify cooperation between two clients through model parameter aggregation. During the cooperation, we incrementally enhance the disparity in their data distributions to promote complementarity. Our investigation explores the alterations in the average accuracy and model similarity with the increase of data complementarity. As depicted in Figure~\ref{fig:intro}, cooperation between the two clients with the highest model similarity does not yield the maximum gains, while cooperation between clients when the data exhibits moderate complementarity is more advantageous, even if their models are not the most similar. Additionally, excessive data complementarity might indicate significant discrepancies in data distributions, rendering the cooperation less effective. The experimental results demonstrate the indispensability of complementarity in the cooperation of FL. Therefore, an intriguing question emerges: \emph{how can we deduce the cooperation gain network among clients by simultaneously considering similarity and complementarity, thereby facilitating more optimal model cooperative learning?}

We present an answer to this question grounded in a thorough analysis of statistical heterogeneity. Briefly, suppose we use $p_{i}(x, y) = p_{i}(x) p_{i}(y | x)$ to denote the joint distribution of the feature $x$ and label $y$ in the $i^{th}$ client. By controlling one of the distributions, it is observed that varied $p(y|x)$ signals the presence of a \emph{concept shift} among clients. A substantial concept shift can detrimentally affect model learning. On the other hand, the limited nature of data within each client makes it challenging to precisely characterize the true local distribution. Hence, varied $p(x)$ which indicates the presence of a \emph{covariate shift} could be beneficial, potentially providing more information gain. Experiments in Figure~\ref{fig:intro} also explain that a moderate covariate shift could introduce complementarity in model learning, leading to enhanced performance. Consequently, we argue that allowing moderate variations in the marginal distribution $p(x)$ while ensuring consistency in the conditional distribution $p(y|x)$ presents a more effective cooperation than merely relying on singular model similarity metrics in previous research. 

Within the aforementioned analysis, we propose a novel cooperation framework by balancing \underline{S}imilarity \underline{a}nd \underline{C}omplementarity, named \texttt{FedSaC}. Specifically, we introduce a cooperation network where each node signifies a client and edges reflect cooperation strength. This network is dynamically optimized, balancing model similarity with feature complementarity. The edge weights denote this balance, ensuring that clients not only collaborate with similar models but also leverage complementary feature insights. We applied the cooperation network to a FL framework, dividing it into two processes: server-side and client-side, achieving personalized interactive cooperation under privacy protection conditions. Leveraging the refined approach, our \texttt{FedSaC} adeptly accommodates  various levels of data heterogeneity and multi-modal scenarios, and effectively identifies the optimal collaborators for each client. 

Our experiments validate the efficacy of \texttt{FedSaC}, demonstrating its ability to consider both similarity and complementarity in cooperation while maintaining a balance. Thanks to this property of \texttt{FedSaC}, it outperforms 12 unimodal and 4 multimodal baselines across various benchmark datasets. Consequently, we conclude that complementarity is indeed beneficial in FL cooperation, rather than solely focusing on similarity.

We summarize our contributions as follows:
\vspace{-.2cm}
\begin{itemize}
    \item We challenge a widely accepted notion that model similarity can be a robust metric for determining the potential benefits of cooperative model learning. We argue that achieving optimal cooperation necessitates a dual consideration of similarity and complementarity.
    \item We propose a novel collaboration framework, \texttt{FedSaC}, which infers the cooperation by optimizing a constrained objective. The objective quantifies a balanced similarity and complementarity between local clients.
    \item We demonstrate through extensive experiments that \texttt{FedSaC} exhibits superior performance in addressing data heterogeneity in FL, surpassing other state-of-the-art FL methods in both unimodal and multimodal scenarios. Code is accessible at \url{https://anonymous.4open.science/r/FedSaC-CE22}
\end{itemize}

\vspace{-.2cm}
\section{Related Work}
\subsection{Federated Learning and Statistical Heterogeneity}
\vspace{-.1cm}

Federated learning~\cite{DBLP:conf/aistats/McMahanMRHA17} has become a key focus in the machine learning field for its practical applications, but it also presents several challenges, including communication efficiency~\cite{DBLP:journals/corr/KonecnyMYRSB16}, privacy concerns~\cite{DBLP:conf/nips/AgarwalSYKM18, DBLP:journals/fgcs/MothukuriPPHDS21}, and statistical heterogeneity~\cite{DBLP:conf/icml/KarimireddyKMRS20, DBLP:conf/kdd/CuiLPCZW22,DBLP:conf/cvpr/QuZLXW00R22}, and they have been the topic of multiple research efforts~\cite{DBLP:conf/icml/MohriSS19}. Recently, a wealth of work has been proposed to handle statistical heterogeneity. For example, ~\cite{DBLP:conf/icml/MohriSS19} seek a balanced model performance distribution by maximizing the model performance on any arbitrary target distribution. ~\cite{DBLP:conf/cvpr/LiHS21} develop MOON that corrects local training by maximizing the similarity between local and global models. Some clustering-based FL methods~\cite{DBLP:journals/corr/abs-2102-04925, DBLP:conf/icml/BaekJJYH23} also attempt to utilize model similarity to cluster similar clients in order to mitigate the impact of statistical heterogeneity. A fundamental question arises ``whether a high degree of model similarity invariably leads to more effective collaboration''?
\vspace{-.1cm}
\subsection{Personalized Federated Learning}
\vspace{-.1cm}
A global model (e.g., FedAvg~\cite{DBLP:conf/aistats/McMahanMRHA17}) could harm certain clients when there are severe distribution discrepancies~\citep{DBLP:journals/corr/abs-2003-13461}, and this stimulates the study of personalized federated learning~\cite{DBLP:conf/cvpr/QuLHDSC23, DBLP:conf/cvpr/QinYWHH23, DBLP:conf/cvpr/0002MB23}. One line of work focused on a better balance between global and local training. For example, there are researches~\citep{DBLP:conf/nips/DinhTN20,DBLP:conf/mlsys/LiSZSTS20,DBLP:conf/icml/KarimireddyKMRS20} proposing to stabilize local training by regulating the deviation from the global model over the parameter space. Another line of research aims to achieve a more fine-grained cooperation via collaboratively learning with similar clients. For example, ~\cite{DBLP:journals/tit/GhoshCYR22} proposed to cluster the collaborative clients according to their model parameter similarity, and learn a personalized model for each cluster. ~\cite{DBLP:conf/icml/YeNWCW23} specify who to collaborate at what intensity level for each client according to model similarity.While the current trend in personalized federated learning heavily relies on similarity metrics, we suggest that a balanced focus on both similarity and complementarity can more accurately optimize collaboration benefits.
\vspace{-.1cm}
\subsection{Federated Multimodal Learning}
\vspace{-.1cm}
Considering the diverse data modalities in real life, there are a few research investigating the tasks of \emph{federated multi-modal learning}, i.e., collaboratively learning models on distributed sources containing multimodal data~\cite{DBLP:journals/ijon/XiongYQX22, DBLP:journals/corr/abs-2109-04833}. In particular, Xiong \emph{et al.} propose a co-attention mechanism\cite{DBLP:journals/ijon/XiongYQX22}~to fuse different modalities. \cite{DBLP:conf/iclr/YuLWXL23}~design a regularization technique to restrict global-local discrepancy by contrastive learning. \cite{DBLP:conf/iotdi/ZhaoBH22}~enhanced the challenge for multimodal clients with unlabeled local data using a semi-supervised framework. Our approach optimizes a weighted sum of model similarity and feature complementarity for automatic weight allocation to clients. Given the higher statistical heterogeneity in multimodal data compared to unimodal data, global models, as currently developed, struggle with conflicting client dependencies. Therefore, our focus is on developing personalized models for multimodal tasks to address data heterogeneity effectively.

\vspace{-.2cm}
\section{Problem Setup}
\vspace{-.1cm}
\label{sec:problem}
The problem to be solved in this paper is formally defined in this section. Specifically, we introduce the objective of federated learning, and through analyzing the statistical heterogeneity,  demonstrate the feasibility and significance of balancing similarity and complementarity in FL.
\vspace{-.1cm}
\subsection{Notations}
\vspace{-.1cm}
Suppose there are $N$ clients in a federated network, each client owns a private dataset $D^{i}$ with $n^{i}$ data samples, where $i = 1,\dots,N$. We define the relative size of each dataset $D^i$ as $p^i = n^i/\sum_{j}n^j$. The dataset $D^{i} = \left\{X^{i}, Y^{i}\right\}$ consists of the input space $X^{i}$ and output space $Y^{i}$.  A data point is denoted by $\left\{x, y\right\}$, with $x$  signifying either a unimodal or a multimodal feature. The input space and the output space are shared across all clients. 

\noindent \textbf{Federated Learning.} In FL scenario, each client collaboratively refines a predictive model using local data and collective knowledge to optimally predict label $y$. 
\texttt{FedAvg}~\cite{DBLP:conf/aistats/McMahanMRHA17}, as an exemplar method introduced by Mcmahan \emph{et al.}, learns a global model $\boldsymbol{\theta}^g$ for all clients by minimizing the empirical risk over the samples from all clients, i.e.,
\begin{equation}
\vspace{-.1cm}
\min _{\boldsymbol{\theta}^g \in \boldsymbol{\Theta}} \  \sum_{i=1}^{N} p^i \mathcal{L}_i\left(\boldsymbol{\theta}^g; D^i\right),
\label{eq:fedavg}
\end{equation}
where $\boldsymbol{\Theta}$ is the hypothesis space and $\mathcal{L}_i$ denotes the loss objective of each clients. From Eq.\ref{eq:fedavg}, \texttt{FedAvg} presumes that i.i.d. data from different clients converge to a shared joint distribution $p(x, y)$, indicating statistical homogeneity across diverse data points.

\vspace{-.1cm}
\subsection{Statistical Heterogeneity}
\vspace{-.1cm}
\label{sec:3-2}
In practical scenarios, the i.i.d assumption in \texttt{FedAvg} is largely unrealistic.
There could be noticeable distinctive traits in local datasets across different clients stemming from diverse environments and contexts in which clients gather data~\cite{DBLP:conf/icml/MohriSS19}. Existing research reveals that such statistical heterogeneity may result in under-performance of global models~\citep{DBLP:conf/cvpr/QuZLXW00R22}.
In the given context, the concept of personalized federated learning is introduced as a potential solution to mitigate the statistical heterogeneity issues by facilitating selective cooperation~\citep{DBLP:conf/icml/00050BS21, DBLP:conf/nips/LinHLZ22, DBLP:conf/icml/YeNWCW23}. Existing works assume that clients derive more benefit from collaborating with peers who possess similar characteristics, thereby implying a diminished level of cooperation where dissimilarities exist. This allows each client to utilize information more akin to their local distribution. However, we pose a fundamental question: \emph{is cooperation with similar peers truly optimal?}

We endeavor to delve deeply into statistical heterogeneity to provide an unexpected answer. Suppose we use $p(x, y)$ to denote the joint distribution of features and labels, the nature of statistical heterogeneity lies in the disparate joint distributions across various clients, i.e., $p(x^{k_1}, y^{k_1}) \not= p(x^{k_2}, y^{k_2})$, where $k_1 \not= k_2$. The joint distribution $p(x, y)$ can be decomposed as $p(x, y) = p(x)p(y|x)$, thus allowing statistical heterogeneity to be represented as 
\begin{equation}
p(x^{k_1})p(y^{k_1} | x^{k_1}) \not= p(x^{k_2})p(y^{k_2} | x^{k_2}),
\label{eq:hetero}
\end{equation}
where $k_1 \not= k_2$. Within this formulation, we recall the following definition.
\begin{definition}
Here is the definition of two distribution shifts.\\
(1) \textbf{\emph{Covarient shift}}~\cite{DBLP:conf/iclr/PengHZS20, DBLP:journals/corr/abs-2112-13381}: The distribution of input features $p(x)$ exhibits disparities among different clients, while the conditional distribution $p(y|x)$ is shared. \\
\noindent (2) \textbf{\emph{Concept shift}}~\cite{DBLP:conf/aistats/JothimurugesanH23, DBLP:conf/ijcnn/CanonacoBMR21}: The relationship between input features and output labels $p(y|x)$ alterations among different clients, even if the distribution of input features $p(x)$ remains constant.
\end{definition}
\vspace{-.1cm}
\subsection{Optimal Cooperation}
\vspace{-.1cm}
To effectively mitigate the challenges of statistical heterogeneity in federated learning, previous research has predominantly focused on employing a similarity metric to facilitate client cooperation. This approach, as highlighted in studies such as ~\citep{DBLP:conf/icml/00050BS21, DBLP:journals/tnn/SattlerMS21, DBLP:conf/icml/YeNWCW23}, emphasizes maximizing similarity to address \emph{concept shift}, which is indeed a crucial aspect of aligning learning models across diverse clients.

\noindent \textbf{Maximizing similarity is justifiable for addressing \emph{concept shift.}} Specifically, when two clients exhibit similar conditional distributions $p(y|x)$, it signifies a shared correlation or mapping between input features and output labels. Such a correlation is instrumental in fostering a more coherent alignment and synthesis of the learned models or knowledge, thereby enhancing the overall effectiveness of the federated learning process. However, in the case of covariant shift, where there is a variation in input features across clients, this strategy may not yield the same level of effectiveness.

\noindent \textbf{Moderate covariant shift is beneficial for cooperation in federated learning}, as exhibited in Figure~\ref{fig:intro}. In the context of federated learning, local clients, limited by their specific data subsets, often present an incomplete representation of the broader data distribution. Therefore, clients are inclined to engage in cooperation to surmount the limitations posed by their individual data paucity. 
When there is minimal covariate shift between two clients, overlapping input features can limit the model's capacity to absorb diverse information. This constraint impedes the detection of underlying data patterns, hindering collaborative efforts. Thus, we suggest that a client should collaborate with peers whose input features exhibit moderate variations. Such strategic collaborations harness complementary data, enhancing the model's predictive accuracy and generalization potential.

Given the aforementioned analysis, it becomes apparent that reliance on similarity metrics as a collaborative criterion is suboptimal in the presence of covariate shift. It would be more judicious to permit moderate variations  in $p(x)$ while maintaining similarity in the conditional distribution $p(y|x)$.

\vspace{-.2cm}
\section{Our Method: Balancing Similarity and Complementarity}
\label{sec:method}

In this section, we detail the construction of a cooperation network, designed to identify optimal collaborators for each client within FL framework. Section~\ref{sec: 4.1} introduces our cooperation network, and Section~\ref{sec: 4.2} discusses the global optimization process, balancing similarity and complementarity. In Section~\ref{sec: 4.3}, we provide specific methods for computing similarity and complementarity under privacy protection, and further decompose the global optimization into server-side and client-side to fit FL architecture.

\begin{figure}[tbp]
\vspace{-.1cm}
\setlength{\abovecaptionskip}{-.1cm}
\setlength{\belowcaptionskip}{-.3cm}
\centering{
\includegraphics[width=0.9\columnwidth]{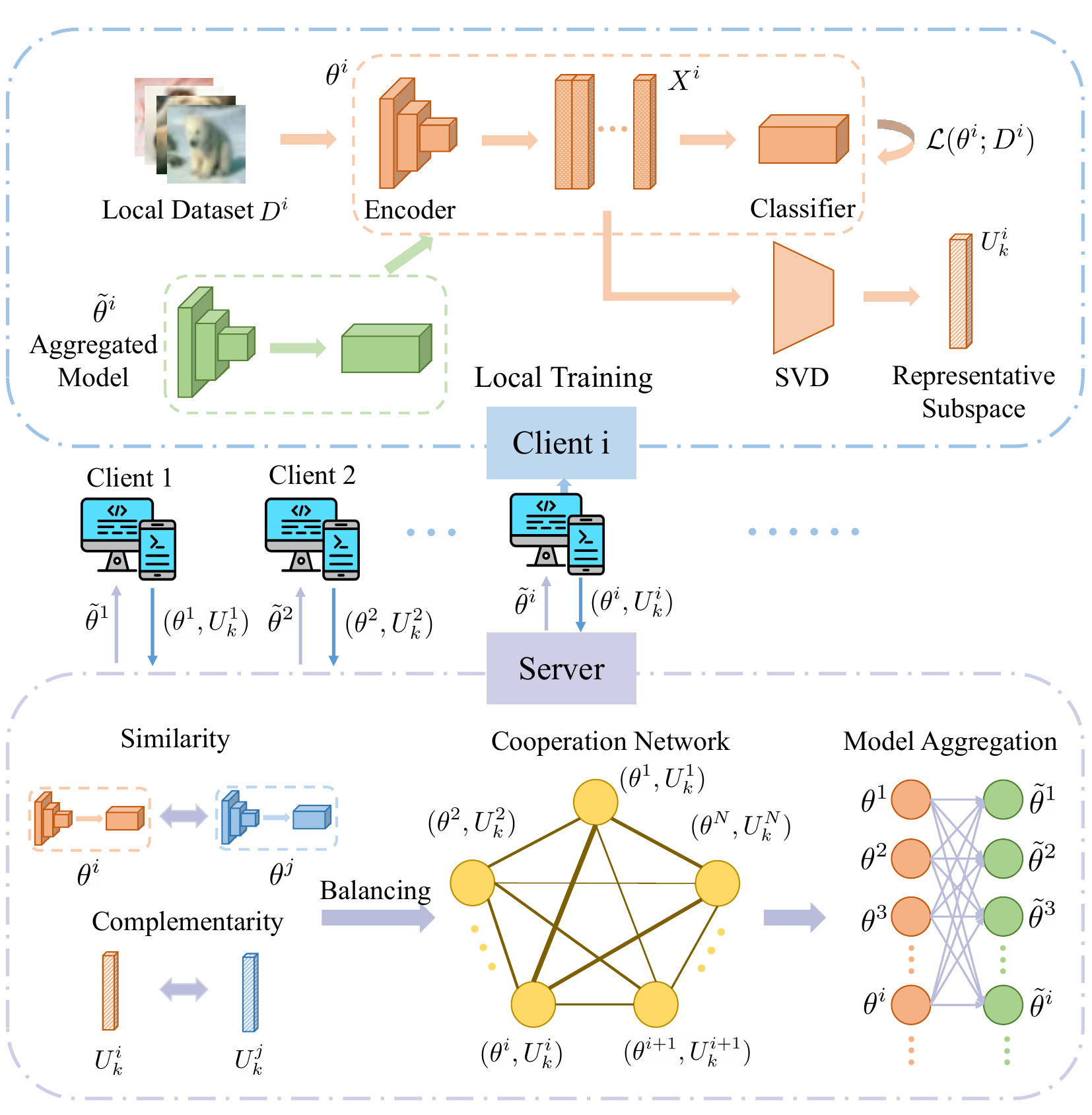}}
\vspace{-.2cm}
\caption{
Illustration of our \texttt{FedSaC} approach. Local clients train models by minimizing empirical risk, incorporating a regularization term based on the distance to the aggregated model. Post-training, models are distilled via SVD to capture the representative subspace, which, alongside model parameters, is sent to the server. The server constructs a cooperation network, balancing similarity and complementarity among clients, to aggregate models. These aggregated models are then disseminated to clients for the subsequent training iteration.
}
\label{fig:method}
\vspace{-.4cm}
\end{figure}
\vspace{-.1cm}
\subsection{Cooperation Network}
\vspace{-.1cm}
\label{sec: 4.1}
Contrary to prior personalized FL works, we argue that merely targeting similar clients is not always optimal, as diverse feature distributions can yield more insights for robust generalization. Hence, our goal of cooperation among federated clients is to achieve a balance: seeking data with a similar conditional distribution while ensuring a complementary marginal distribution.

To measure this balance, we introduce two metrics: similarity, targeting minimal concept shifts within similar conditional distributions, and complementarity, addressing moderate covariate shifts for diverse marginal distributions. Informed by this rationale as discussed in Section~\ref{sec:3-2}, we advocate that clients should collaborate with others who share similar $p(y|x)$ but different $p(x)$. Such a balanced collaboration ensures that clients not only access shared knowledge but also harness complementary insights from different data angles, ultimately boosting learning outcomes.

From a global perspective, the cooperation strength of a client is influenced by other clients. Inspired by~\cite{DBLP:conf/icml/YeNWCW23}, we construct a cooperation network shown in Figure~\ref{fig:method} which balances similarity and complementarity among clients while collaborating. This network comprises $N$ nodes, each representing a client. The adjacency matrix of the network is denoted as $\boldsymbol{W} \in \mathbb{R}^{N\times N}$, where the element $\boldsymbol{W}_{ij}$ indicates the cooperation strength between the $i^{th}$ and $j^{th}$ clients in federated learning. We establish a global objective encompassing both similarity and complementarity to determine the optimal weights in the adjacency matrix, which identifies the optimal collaborators for each client.
\vspace{-.1cm}
\subsection{Optimization with Similarity and Complementarity}
\vspace{-.1cm}
\label{sec: 4.2}
In practice, client data distributions are inaccessible. To bypass the privacy constraint, we utilize local models as surrogates for estimating data distributions.
We capture the local data distribution of clients as the sampling distribution for their marginal distribution $p(x)$.
For the conditional distribution $p(y|x)$, intuitively, the personalized model parameters, after local training, can capture the mapping from the marginal distribution $p(x)$ to the label distribution $p(y)$. 
Hence, we consider the local-trained parameters as  approximate surrogates for the conditional distribution $p(y|x)$.
 We present a global optimization equation, as articulated in Equation~\ref{eq:global_optimal}, which aims to refine the local personalized model parameters $\{\boldsymbol{\theta}^i\}$ for each client and the network adjacency matrix $\boldsymbol{W}$. The term $\mathcal{L}_{i}(\sum_{j=1}^N\boldsymbol{W}_{ij}\boldsymbol{\theta}^j; D^i)$ denotes the empirical risk on the local dataset of the $i^{th}$ client, following the weighted aggregation
of model parameters across multiple clients. $\mathcal{C}$ denotes the complementarity of marginal distributions between two clients,  while 
$\mathcal{S}$ denotes the similarity in model parameters between them. The hyperparameters $\alpha$ and $\beta$ are introduced to adjust the prominence of complementarity and similarity, ensuring a balanced emphasis on both during optimization.
{
\fontsize{9.5pt}{10pt}
\vspace{-.1cm}
\begin{align}
\min\limits_{\{\boldsymbol{\theta}^i\}, \boldsymbol{W}}&\sum_{i=1}^Np^i\Biggl(\mathcal{L}_{i}(\sum_{j=1}^N\boldsymbol{W}_{ij}\boldsymbol{\theta}^j; D^i) + \alpha\sum_{j=1}^N\mathcal{C}(\boldsymbol{W}_{ij};D^i; D^j)\nonumber \\ & \quad \quad \quad - \beta\sum_{j=1}^N\mathcal{S}(\boldsymbol{W}_{ij};\boldsymbol{\theta}^i; \boldsymbol{\theta}^j)\Biggr)\nonumber
\\ & \text{s.t.} \quad \sum_{j=1}^N\boldsymbol{W}_{ij}=1,\forall{i};\quad \boldsymbol{W}_{ij}\geq 0,\forall{i, j} 
\label{eq:global_optimal}
\end{align}
}

\vspace{-.3cm}
The optimization equation minimizes the empirical risk on local data while balancing the similarity and complementarity among clients. The two constraints ensure the normalization and non-negativity of each client's cooperation weight. Compared to previous methods, our cooperation network approach flexibly determines the cooperation strength among clients. By considering variations in marginal distributions across different clients, it more effectively captures distributional differences, leading to enhanced model performance. 

\vspace{-.2cm}
\subsection{FedSaC: Balancing Similarity and Complementarity}
\vspace{-.1cm}
\label{sec: 4.3}
In practical FL architectures, each client is restricted to its local dataset and model. Model aggregation, as well as the computation of complementarity and similarity, require coordination with a central server. Given this structure, we initially introduce the metric of similarity and complementarity under privacy constraints. Subsequently, we partition the global optimization equation into two stages, optimizing separately at the server side and the client side.
The aforementioned process is illustrated in Figure ~\ref{fig:method}.
{
\vspace{-.1cm}
\begin{algorithm}[htbp]
\caption{FedSac}
\setlength{\belowcaptionskip}{-5pt}
\noindent {\bf Input:} 
Total communication round $T$, client number $N$, initial local model $\{\boldsymbol{\theta}_{(0)}^i\}$, initial  cooperation network $\boldsymbol{W}_{(0)}$;
\begin{algorithmic}[1]
\FOR{each round $t=0,..., T-1$}
    \STATE {\bf Client Side}
    \FOR{client $i=1,..., N$ in parallel}
        \STATE Receive aggregated model $\tilde{\boldsymbol{\theta}}_{(t)}^i$ sent from server;
        \STATE Update local model $\boldsymbol{\theta}_{(t)}^i \leftarrow \tilde{\boldsymbol{\theta}}_{(t)}^i$;
        \STATE Minimize local loss defined in Eq.\ref{eq:client side};
        \STATE Extract representative subspace $U_{k(t)}^i$ by Eq.\ref{eq:SVD}; 
        \STATE Send $\boldsymbol{\theta}_{(t)}^i$ and $U_{k(t)}^i$ to server;
    \ENDFOR
    \STATE {\bf Server Side}
    \STATE Calculate similarity $\mathcal{S}$ for each pair of clients as Eq.\ref{eq:similarity};
    \STATE Calculate complementarity $\mathcal{C}$ for each pair of clients as Eq.\ref{eq:principal} and Eq.\ref{eq:complementarity};
    \STATE Update $\boldsymbol{W}_{(t)}$ by optimization Eq.\ref{eq:server};
    \STATE Aggregate model $\tilde{\boldsymbol{\theta}}_{(t+1)}^i \leftarrow \sum_j\boldsymbol{W}_{ij(t)}\boldsymbol{\theta}_{(t)}^j$;
\ENDFOR
\STATE \textbf{Output:} the learned personalized models $\left\{\boldsymbol{\theta}_{(T)}^{i} \right\}$.
\end{algorithmic}
\label{alg}
\end{algorithm}
\vspace{-.2cm}
}

\vspace{-.2cm}
\subsubsection{The Metric of Similarity and Complementarity}
\vspace{-.1cm}
\noindent \textbf{Similarity Metric.} Following conventional practices, we use model parameters as proxies and adopt the cosine distance between the local models of the $i^{th}$ and $j^{th}$ clients as our similarity metric, denoted as,
\vspace{-.1cm}
\begin{equation}
\vspace{-.1cm}
\mathcal{S}(\boldsymbol{W}_{ij};\boldsymbol{\theta}^i; \boldsymbol{\theta}^j)=\boldsymbol{W}_{ij}\frac{\boldsymbol{\theta}^i\cdot\boldsymbol{\theta}^j}{\|\boldsymbol{\theta}^i\|\cdot\|\boldsymbol{\theta}^j\|}.
\label{eq:similarity}
\vspace{-.1cm}
\end{equation}

\noindent \textbf{Complementarity Metric.} In light of the privacy principles inherent to FL, we use an indirect method to capture data complementarity. For a given local dataset $D^i$ at client $i$,  the local model $\boldsymbol{\theta}^i$ extracts the feature matrix $\boldsymbol{X}^i$. Applying singular value decomposition(SVD) on $\boldsymbol{X}^i$ yields:
\begin{equation}
\vspace{-.1cm}
\boldsymbol{X}^i = \boldsymbol{U}^i\boldsymbol{\Sigma}^i(\boldsymbol{V}^i)^T,
\label{eq:SVD}
\vspace{-.1cm}
\end{equation}
where $\boldsymbol{U}^i$ contains the singular vectors of $\boldsymbol{X}^i$, capturing the direction in the feature space. For our purposes, we consider the first $k$ columns of $\boldsymbol{U}^i$, denoted $\boldsymbol{U}^i_k$, as the representative subspace for client $i$.

To gauge the complementarity between clients $i$ and $j$, we utilize the principal angles between their respective subspaces. The $l^{th}$ principal angle $\cos\phi_l$ between the two is given by:
\begin{equation}
\cos\phi_l = \max\nolimits_{u \in \boldsymbol{U}^i_k, v \in \boldsymbol{U}^j_k}u^Tv,
\label{eq:principal}
\end{equation}
where $l=1,\dots, k$. These angles offer a quantifiable measure of the complementarity between the two datasets. A small principal angle suggests a high similarity between the subspaces, while an angle close to $\pi/2$ implies that the subspaces are nearly orthogonal, indicating significant divergence in their feature spaces.

By averaging these angles, we obtain complementarity as:
\begin{equation}
\vspace{-.1cm}
\mathcal{C}(\boldsymbol{W}_{ij};D^i; D^j)=\boldsymbol{W}_{ij}\cos\left(\frac{1}{k}\cdot\sum_l\phi_l\right).
\label{eq:complementarity}
\vspace{-.1cm}
\end{equation}
\vspace{-.2cm}
\subsubsection{FedSaC in FL architecture}
\vspace{-.1cm}
\noindent \textbf{Server Side.}
On the server side, we compute the similarity and complementarity based on the model parameters $\{\boldsymbol{\theta}^i\}$ and the subspace representation $\{\boldsymbol{U}^i_k\}$ received from the local client. Subsequently, we derive the adjacency matrix $\boldsymbol{W}$ through optimization equations. In FL scenario, as the empirical loss of local clients is elusive, we utilize the relative dataset size $p_i$ as a surrogate measure. Clients with larger datasets are considered more reliable collaborators and should thus be assigned greater cooperative weight. Given these considerations, the optimization equation on the server side is defined as:
{
\vspace{-.1cm}
\fontsize{9.2pt}{10pt}
\begin{align}
\min\limits_{\boldsymbol{W}_{i*}}&\sum_{j=1}^N\Biggl((\boldsymbol{W}_{ij}-p^j)^2 + \alpha\mathcal{C}(\boldsymbol{W}_{ij};D^i; D^j) - \beta\mathcal{S}(\boldsymbol{W}_{ij};\boldsymbol{\theta}^i; \boldsymbol{\theta}^j)\Biggr)\nonumber
\\ & \text{s.t.} \quad \quad \quad \sum_{j=1}^N\boldsymbol{W}_{ij}=1,\forall{i};\quad \boldsymbol{W}_{ij}\geq 0,\forall{i, j} 
\label{eq:server}
\end{align}
\vspace{-.4cm}
}

Using the cooperation network $\boldsymbol{W}$, we derive the aggregated model $\tilde{\boldsymbol{\theta}}^i = \sum_j\boldsymbol{W}_{ij}\boldsymbol{\theta}^j $ for each client.

\noindent \textbf{Client Side.}
On each client side, our objective is to minimize the local empirical risk while preventing overfitting of the aggregated model on the local dataset. We replace the current local model with the aggregated model received from the server $\boldsymbol{\theta}^i \leftarrow \tilde{\boldsymbol{\theta}}^i$,  and further refine this local model. For the $i^{th}$ client, the optimization equation is defined as:
\begin{equation}
\vspace{-.1cm}
\arg\min\limits_{\boldsymbol{\theta}^i}\  
\mathcal{L}_{i}(\boldsymbol{\theta}^i; D^i) - \lambda\cos(\boldsymbol{\theta}^i, \tilde{\boldsymbol{\theta}}^i),
\label{eq:client side}
\vspace{-.1cm}
\end{equation}
where $\cos(\boldsymbol{\theta}^i, \tilde{\boldsymbol{\theta}}^i)$  ensures that the locally optimized model does not deviate excessively from the aggregated model, and $\lambda$ represents the regularization hyperparameter. The optimized model then serves as the current local model for participation in the subsequent optimization round.

\vspace{-.2cm}
\section{Experiments}
\label{sec:experiments}

\subsection{Unimodal Experiments Setup}
\begin{table*}
\vspace{-.1cm}
\centering
\begin{tabular}{c|c c c c|c c c c}
\hline
\multirow{2}{*}{\begin{tabular}{@{}c@{}}Dataset\\H-Level\end{tabular}} & \multicolumn{4}{c|}{CIFAR-10} & \multicolumn{4}{c}{CIFAR-100} \\
\cline{2-9}
& Homo & Diri(low) & Diri(high) & Pathol & Homo & Diri(low) & Diri(high) & Pathol\\
\hline
Local &54.81 & 72.72 &83.83 &91.07 & 16.13 & 30.63 & 47.68 & 49.42  \\
FedAvg & 67.12& 63.61 & 62.92 & 66.19 & 31.10 & 30.66 & 27.78 & 26.23 \\
FedProx  &62.92 & 62.93 & 62.25 & 55.76 & 30.55 & 30.64 & 27.87 & 25.64 \\
CFL & 60.55 & 73.81& 83.84 & 90.76 & 19.31 & 33.21 & 49.12 & 52.43 \\
pFedMe & 47.48 & 66.35 &75.24 & 81.73 & 13.18 & 25.18 & 34.37 & 33.48 \\
Ditto & 65.35 & 75.98 & 83.78 & 89.41 & 29.41 & 39.73 & 50.33 & 50.54 \\
FedAMP & 45.49 & 64.29 &75.49 & 86.90 & 10.07 & 22.66 & 31.04 & 37.50 \\
FedRep & 62.88 & 74.14 &83.47 & 90.02 & 21.53 & 34.72 & 50.15 & 26.23 \\
pFedHN & 62.78 & 66.62 & 82.57 & 89.91 & 25.94 & 30.89 & 49.08 & 49.06 \\
FedRoD & 62.07 &74.06 & 83.49 & 90.66 & 18.71 & 31.65 & 47.96 & 49.91 \\
kNN-Per &67.01 & 63.05 &70.05 & 79.09 & 31.04 & 30.95 & 25.84 & 24.70 \\
pFedGraph &67.37 & 75.22 & 84.28 & 92.74& 31.16 & 38.71 & 51.63 & 56.79 \\
\textbf{FedSaC} &\textbf{70.89} & \textbf{80.46} & \textbf{93.14} & \textbf{92.89} & \textbf{34.84} & \textbf{41.80} & \textbf{56.27} & \textbf{57.48}\\ 
\hline
\end{tabular}
\vspace{-.2cm}
\caption{Average accuracy of our unimodal FedSaC on CIFAR-10 and CIFAR-100 dataset}
\label{tab:unimodal}
\vspace{-.3cm}
\end{table*}

\noindent \textbf{Datasets and Data Heterogeneity.} Following the predominant experimental setup in personalized federated learning, we evaluate our proposed \texttt{FedSaC} on two image classification datasets:  CIFAR-10 and CIFAR-100~\cite{krizhevsky2009learning}. For each dataset, we implement four partitions with different heterogeneous levels into $K$ clients~\cite{DBLP:conf/icml/YeNWCW23}. 1) Homogeneous partition, where each client is imbued with data samples under a uniform probability schema. 2) Dirichlet partition~\cite{DBLP:conf/icml/YurochkinAGGHK19}, where the allocation ratio of data samples from each category is instantiated from $Dir_K(\alpha)$.  Notably, we define heterogeneity levels with $\alpha=0.1$ (high) and $\alpha=0.5$ (low). 3) Pathological partition, where each client is assigned with data exclusively from 2 categories for 10 classification datasets and 20 categories for 100 classification datasets.

\noindent \textbf{Baselines.} We compare our \texttt{FedSaC} with 12 representative FL approaches including: 1) \textbf{Local}: Local training without information sharing. 2) \emph{FedAvg}~\cite{DBLP:conf/aistats/McMahanMRHA17} and 3) \emph{FedProx}~\cite{DBLP:journals/corr/abs-2003-13461}: Popular FL methods where local updates are centrally aggregated. 4) \emph{CFL}~\cite{DBLP:journals/tnn/SattlerMS21}: Clustered FL for client group learning. 5) \emph{pFedMe}~\cite{DBLP:conf/nips/0001MO20}: Using regularized loss functions to decouple local and global models. 6) \emph{Ditto}~\cite{DBLP:conf/icml/00050BS21}: Enhanced robustness and fairness by regularized optimization. 7) \emph{FedAMP}~\cite{DBLP:conf/aaai/HuangCZWLPZ21}:  Pairwise collaboration between similar clients in FL. 8) \emph{FedRep}~\cite{DBLP:conf/icml/CollinsHMS21}: Shared data representation with local client heads. 9) \emph{pFedHN}~\cite{DBLP:conf/icml/ShamsianNFC21}: Hypernetworks generate unique client models in personalized FL. 10) \emph{FedRoD}~\cite{DBLP:conf/iclr/ChenC22}: Decoupled framework balances generic and personalized predictors 11) \emph{kNN-Per}~\cite{DBLP:journals/corr/abs-2111-09360}: Personalization via global embeddings and local kNN interpolation. 12)\emph{pFedGraph}~\cite{DBLP:conf/icml/YeNWCW23}: Adaptive collaboration via learned graph.
\vspace{-.2cm}


\subsection{Unimodal Experimental Results}
\vspace{-.1cm}
Our experimental evaluations, conducted across various levels of heterogeneity on the CIFAR-10 and CIFAR-100 datasets, conclusively demonstrate the superior performance of our proposed model, \texttt{FedSaC}. Our analysis of the results presented in Table~\ref{tab:unimodal} leads to two primary insights: 

\noindent \textbf{Superiority Across Heterogeneity Levels.}  \texttt{FedSaC} consistently surpasses baseline models in various heterogeneity settings, highlighting its superior performance. Experiments show that our personalized method notably outperforms conventional techniques like FedAvg, especially in situations with statistical heterogeneity.  Furthermore,  \texttt{FedSaC} demonstrates significant or comparable enhancements over pFedGraph, a method based on similarity metrics. This comparison emphasizes the efficacy of our balanced approach to similarity and complementarity, a crucial aspect in federated learning for handling diverse data distributions.

\noindent \textbf{Enhanced Performance under Strong Complementarity.} \texttt{FedSaC} excels in scenarios with significant complementarity, such as those involving Dirichlet partitioning. These settings often feature imbalanced data distributions, posing challenges for local models with limited data categories. \texttt{FedSaC}'s effective balance of similarity and complementarity addresses these challenges, enhancing data representation. In Dirichlet partitioning, it consistently surpasses pFedGraph, which relies on similarity metrics, improving accuracy by 3\% to 8\%. This alignment of empirical results with our theoretical framework confirms the effectiveness and versatility of \texttt{FedSaC} in various federated learning contexts, especially with high data heterogeneity.

\vspace{-.2cm}
\subsection{Multimodal Experiments Setup}
\vspace{-.1cm}
\begin{table*}
\centering
\begin{tabular}{c|c c | c c |c c | c c}
\hline
\multirow{2}{*}{\begin{tabular}{@{}c@{}}H-Level\\ modal \end{tabular}} & \multicolumn{2}{c|}{Homo} & \multicolumn{2}{c|}{Diri(low)} & \multicolumn{2}{c|}{Diri(high)} & \multicolumn{2}{c}{Pathol}\\
\cline{2-9}
&  visual & textual & visual & textual & visual & textual & visual & textual  \\
\hline
Local & 9.42 & 9.65 & 18.45 &17.01 & 28.46 & 24.44 & 19.81 & 17.12 \\
FedAvg & 20.13 & 16.97 & 19.81 & 16.48 & 21.41 & 16.01 & 19.80 & 16.65\\
FedIoT  & 19.96 & 17.24 & 19.92 & 17.45 & 20.04 & 17.23 & 19.69 & 16.95\\
pFedGraph  &21.87 & 21.59 & 24.48 & 25.13 & 31.78 & 31.34 & 25.50& 27.06\\
\textbf{FedSaC} &\textbf{25.25} & \textbf{22.97} & \textbf{28.62} & \textbf{27.40} & \textbf{35.36} & \textbf{33.44} & \textbf{30.07} & \textbf{27.52}\\ 
\hline
\end{tabular}
\vspace{-.2cm}
\caption{Average accuracy of our multimodal FedSaC on CUB200-2011 dataset}
\label{tab:multimodal}
\setlength{\belowcaptionskip}{-4pt}
\vspace{-.3cm}
\end{table*}

\noindent \textbf{Datasets and Baselines.} In our multimodal experiments, we employ the \emph{CUB200-2011}~\cite{welinder2010caltech} multimodal dataset, which encompasses two modalities—images and text—to undertake the task of classifying 200 bird species. 
For multimodal baselines, we not only compare with local training but also extend unimodal methods \emph{FedAvg} and \emph{pFedGraph} to the multimodal context, executing tasks separately within each modality. Additionally, we incorporate the multimodal federated learning method \emph{FedIoT}~\cite{DBLP:conf/iotdi/ZhaoBH22} for comparison. This method conducts unsupervised training on local clients and supervised aggregation on the server.

\noindent \textbf{Multimodal Setup.}  Our proficient unimodal \texttt{FedSaC} method has been expanded to multimodal experimentation. Unlike unimodal scenarios, the multimodal approach leverages inter-client complementarity to enhance personalized model performance and utilizes inter-modality complementarity to contribute additional information to the model. Therefore, we introduce a strategy for the fusion of multimodal information complementarity. The specific setup details will be presented in Appendix~\ref{sec:app_b}.


\vspace{-.2cm}
\subsection{Multimodal Experimental Results}
\vspace{-.1cm}
The multimodal experimental results, as depicted in Table~\ref{tab:multimodal}, demonstrate that our \texttt{FedSaC} method surpasses all baselines. It significantly outperforms FedIoT, a method tailored for multimodal federated learning, which validates the efficacy of \texttt{FedSaC} in handling complex multimodal data. Particularly in scenarios modeled by Dirichlet distributions, our method demonstrates a distinct advantage over other baselines, reflecting a consistent trend with our unimodal experiment outcomes. Notably, we observe a more pronounced improvement in the visual modality post-cooperation, suggesting that visual data may provide richer information that enhances the robustness of the \texttt{FedSaC} method's cooperative framework.

\vspace{-.2cm}
\subsection{Visualization}
\vspace{-.1cm}

\begin{figure}[tbp]
\centering
\subfigure[Data Sim]{
    \includegraphics[width=0.29\columnwidth]{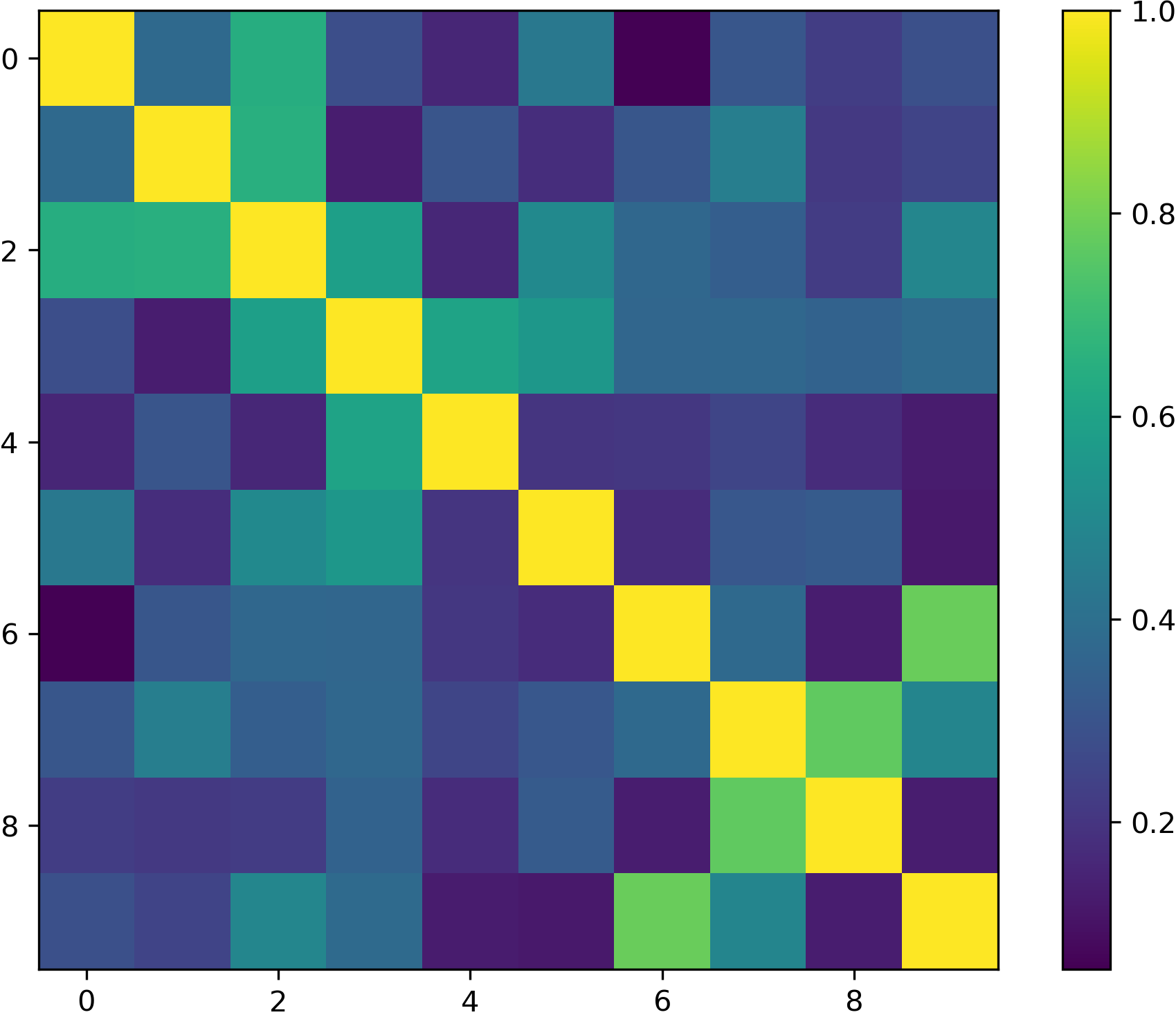}
    \label{fig:a}
}
\hfill 
\subfigure[Model Sim]{
    \includegraphics[width=0.29\columnwidth]{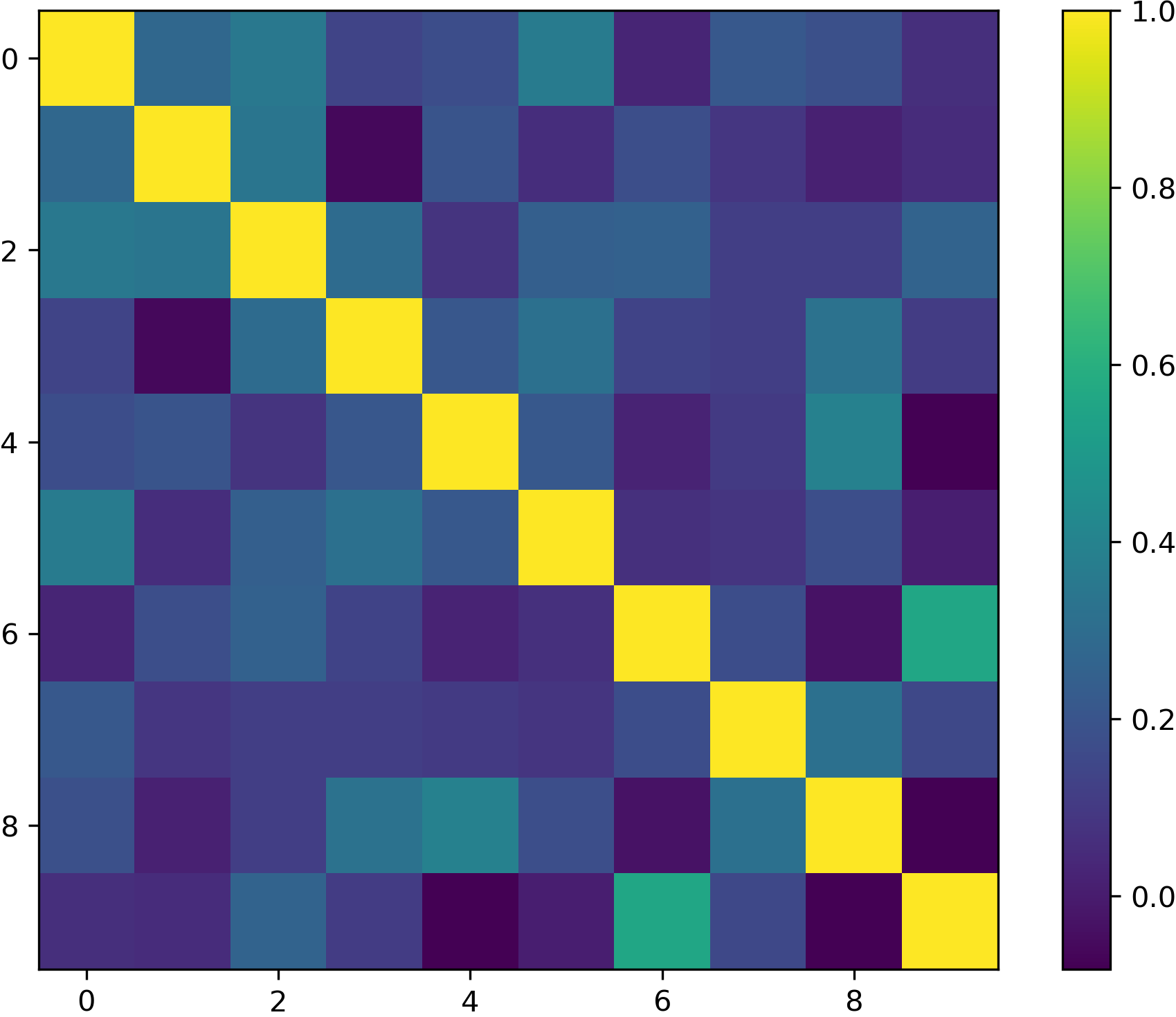}
    \label{fig:b}
}
\hfill 
\subfigure[Feature Comp]{
    \includegraphics[width=0.29\columnwidth]{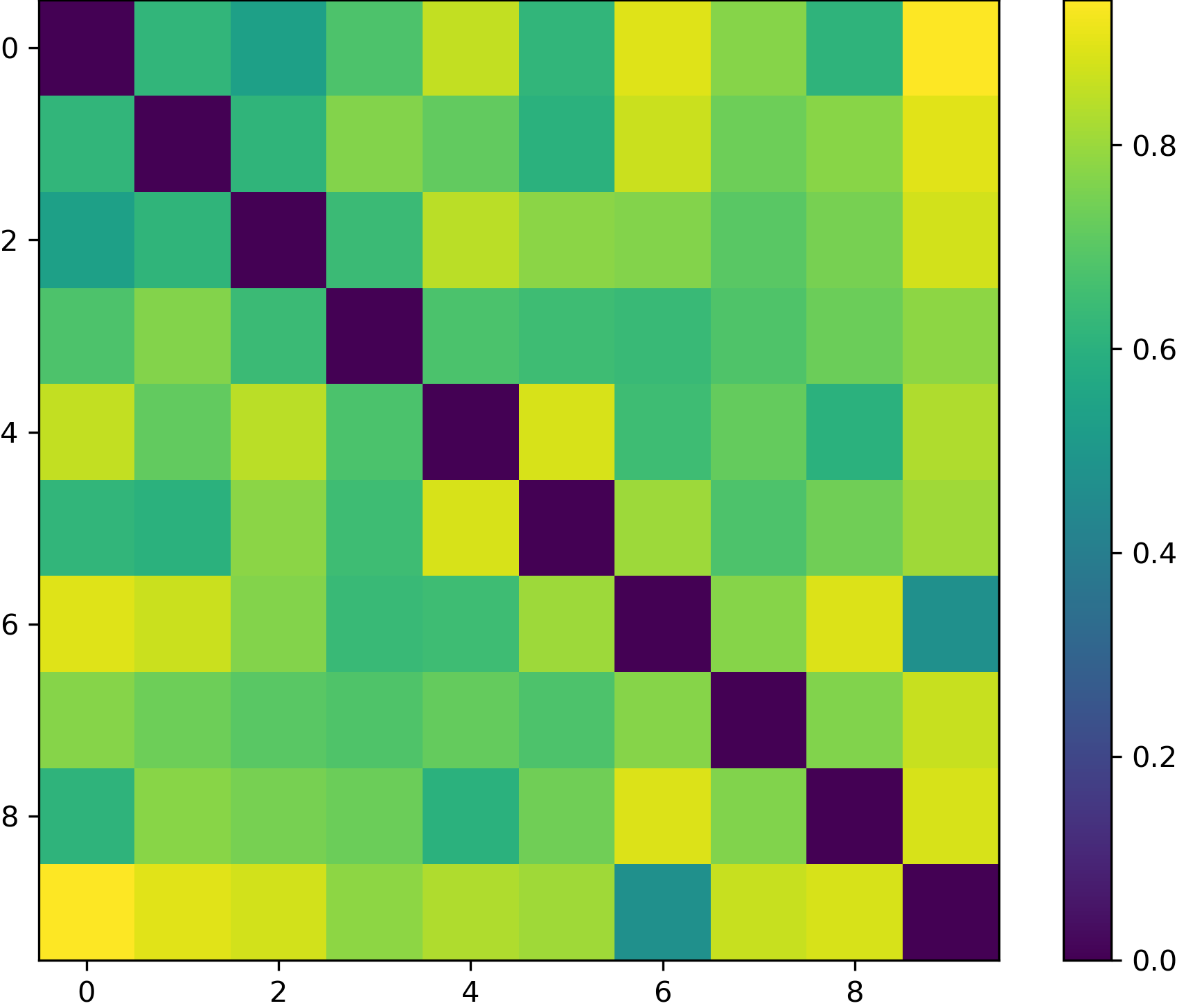}
    \label{fig:c}
}

\subfigure[Sim Network]{
    \includegraphics[width=0.29\columnwidth]{figs/origin_matrix.png}
    \label{fig:d}
}
\hfill 
\subfigure[Comp Network]{
    \includegraphics[width=0.29\columnwidth]{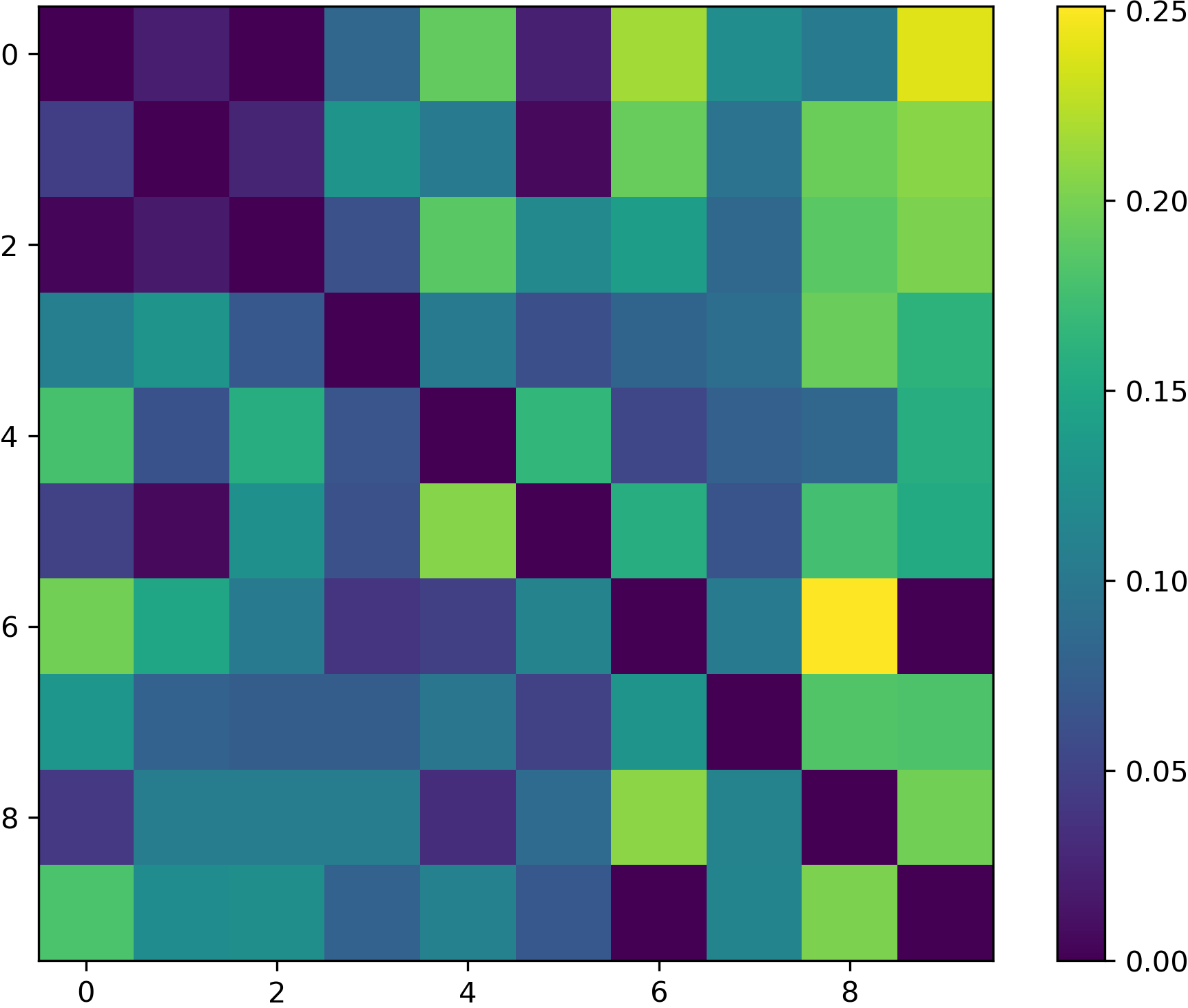}
    \label{fig:e}
}
\hfill 
\subfigure[Bal Network]{
    \includegraphics[width=0.29\columnwidth]{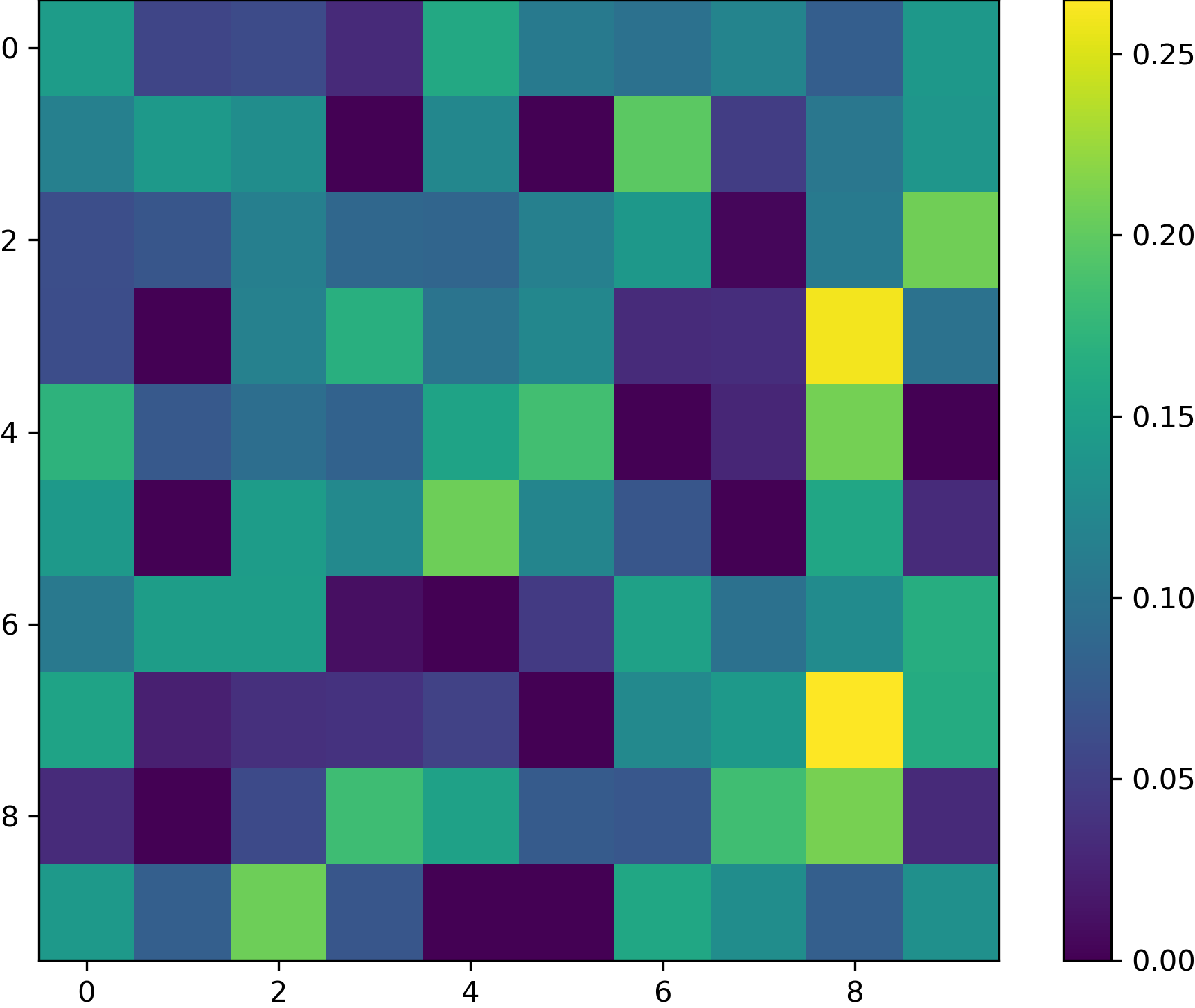}
    \label{fig:f}
}
\vspace{-.3cm}
\caption{Visualization of FedSaC: local data, process matrices, and cooperation networks under three collaboration states}
\label{fig:visual}
\vspace{-.7cm}
\end{figure}

In our \texttt{FedSaC} visualization, Figure~\ref{fig:visual} presents core matrices and cooperation networks. Figure~\ref{fig:a} shows local data's cosine similarity, while Figures~\ref{fig:b} and \ref{fig:c} display the model similarity and feature complementarity matrices, respectively. The comparison of Figures~\ref{fig:a} and \ref{fig:c} demonstrates a complementary pattern, affirming our metric's effectiveness in capturing local data relationships under privacy constraints. Figures~\ref{fig:d} to \ref{fig:f} depict cooperation networks under three collaboration scenarios: focusing on similarity, complementarity, and a balance of both. It is observed that in the similarity-based network, clients predominantly maintain their own models, hindering cooperative effectiveness and information gain.  In the complementarity network, clients almost completely abandon their initial states, which is disadvantageous for training. The balanced approach allows for probabilistic exploration while filtering out clients with excessively high heterogeneity, as indicated by the darker areas that also show inconsistency in the local data matrix. 
The visualization underscores our method's role in boosting FL collaboration efficiency.

\setlength{\belowcaptionskip}{-.1cm}

\vspace{-.2cm}
\subsection{More Discussion }
\vspace{-.1cm}


\textbf{Large-Scale Clients.} In our experiments with a smaller scale of client data, we enhanced cooperation efficiency in large-scale client collaborations (e.g., with 50 or 100 clients) by randomly selecting a subset of clients in each iteration. The feasibility of this approach is demonstrated in Appendix~\ref{sec:large_scale}. 

\textbf{Communication Overhead.} Despite the additional steps introduced for optimal cooperation, Appendix~\ref{sec:compute} analyzes and confirms that the extra Computational cost is minimal compared to local training, and thus acceptable. 

\vspace{-.2cm}
\section{Conclusion}
\label{sec:conclusion}
\vspace{-.1cm}
In this study, we investigate the complex dynamics of federated learning, mitigating the significant challenge of statistical heterogeneity. We shift the focus from model similarity to a balance between similarity and feature complementarity. Our framework, FedSaC, effectively constructs a cooperation network by optimizing this balance. Extensive experiments show FedSaC's superiority over current FL methods in various scenarios. This research challenges conventional approaches and contributes to developing more robust learning models for complex federated settings.
\vspace{-.2cm}
\section{Impact Statement}
\vspace{-.1cm}

This study presents the \texttt{FedSaC} framework, offering a strategic approach to address statistical heterogeneity in Federated Learning. Academically, it introduces a novel perspective to FL, encouraging future research to explore the interplay of complementarity and similarity in model cooperation. Practically, this framework can be flexibly applied across various industries, facilitating more efficient and privacy-preserving data analysis models. Ethically, \texttt{FedSaC} aligns with the increasing demand for ethical data use and user privacy in technological advancements. Future work will further investigate the significance of balancing similarity and complementarity in multimodal architectures.


\nocite{langley00}

\bibliography{example_paper}
\bibliographystyle{icml2024}

\newpage
\appendix
\onecolumn
\section{Discussions about FedSac.}
\label{sec:app_a}
\subsection{FedSac Optimization}
\noindent \textbf{Global Optimization.} In our research, the overarching optimization equation is presented as Equation 3, which is fundamentally grounded in the optimization objective of FedAvg~\cite{DBLP:conf/aistats/McMahanMRHA17}. The equation is expressed as follows:
\begin{equation}
\min _{\{\boldsymbol{\theta}^i\}}  \sum_{i=1}^{N} p^i \mathcal{L}\left(\boldsymbol{\theta}^g; D^i\right),
\end{equation}
This equation is restructured to align with our targeted optimization goals. The global model $\theta_g$ can be represented as a weighted aggregation of local models, with the weights corresponding to the relative sizes of each client's dataset. The revised formulation is presented thusly:
\begin{equation}
\begin{split}
\min\limits_{\{\boldsymbol{\theta}^i\}, \boldsymbol{W}}&\sum_{i=1}^Np^i \mathcal{L}_{i}(\sum_{j=1}^N\boldsymbol{W}_{ij}\boldsymbol{\theta}^j; D^i)
\\ \text{s.t.} \quad \boldsymbol{W}_{ij}=p^j, &\forall{i, j};  \quad \sum_{j=1}^N\boldsymbol{W}_{ij}=1,\forall{i};\quad \boldsymbol{W}_{ij}\geq 0,\forall{i, j} 
\end{split}
\end{equation}
Further to this, we introduced two additional regularization terms to balance similarity and complementarity.  $\mathcal{C}$ objective reduces cooperation intensity between clients with similar datasets, while $\mathcal{S}$ objective increases it for clients with similar model parameters.\\

\noindent \textbf{Server-Client Optimization.} Within the federated learning framework, Equation 3 poses practical challenges, as clients should not directly receive models from other clients. Consequently, transferring models to a centralized server becomes essential. At the server side, we aim to estimate the first term of Equation 3, namely the empirical loss of local models. In line with our optimization objectives, which are aligned with FedAvg, we adopt a FedAvg-inspired approach. Here, we approximate the empirical loss using the relative sizes of the datasets, operating under the premise that clients with larger local datasets are more suitable for collaboration. This concept has been validated for its rationality~\cite{DBLP:conf/icml/YeNWCW23} in federated learning scenarios.

\subsection{The Metric of Similarity and Complementarity}
\noindent \textbf{Similarity Metric.} In the realm of federated learning, utilizing model parameters to gauge client similarity is a prevalent approach~\cite{DBLP:conf/icml/YeNWCW23, DBLP:conf/aaai/HuangCZWLPZ21}. Aligning with the approach in~\cite{DBLP:conf/icml/YeNWCW23}, we utilize cosine distance of model parameters for similarity assessment. \\

\noindent \textbf{Complementarity Metric.} Considering the privacy concerns in federated learning, direct computation of distances using local datasets is not feasible. Instead, we draw upon the principle angle method, a technique that measures distances between subspaces~\cite{miao1992principal}. This adapted approach relies on limited, non-sensitive information to determine the degree of similarity between clients.

The principle angle method offers a geometric perspective for measuring the distance between subspaces. Specifically, when dealing with two subspaces, $\boldsymbol{V}$ and $\boldsymbol{W}$, the method determines the angles $\theta_1, \theta_2, \cdots, \theta_k$ between them. Here, $k$ represents the number of dimensions in the smaller of the two subspaces. The calculation of the $i^{th}$ principal angle is as described in Equation 6. This implies that the cosine of the largest principal angle corresponds to the largest singular value of the matrix product $V^TW$.
\begin{equation}
\cos \theta_i = \frac{<x_i, y_i>}{\|x_i\|\cdot\|y_i\|} = \max\nolimits_{x_i \in \boldsymbol{V}, y_i \in \boldsymbol{W}} \frac{<x_i, y_i>}{\|x_i\|\cdot\|y_i\|}.
\label{priciple}
\end{equation}
The principal angle method provides a clear geometric perspective on how similar or different two subspaces are. By measuring the angles between the subspaces, it offers a more intuitive understanding of their relationship.

In our method, we harness the principal angle method to effectively represent the local data distributions of clients as model output features. This is achieved through SVD, where we select the leading $k$ principal component vectors to represent each client's data in a subspace. This approach maps varied client datasets to a common feature space and ensures privacy by using only a few principal components, which are insufficient to reconstruct the original data. The technique aligns data from different clients effectively while maintaining privacy in federated learning.

\section{Experiments and Implementation Details}
\label{sec:app_b}
\subsection{Unimodal Implementation Details} 
\noindent \textbf{Basic Setup.}Adhering to the training setting presented in \cite{DBLP:conf/icml/YeNWCW23} , we partition the dataset across 10 local clients. Each client utilizes a simple CNN classifier consisting of 2 convolutional layers, 2 subsequent fully-connected layers and a final classification layer. Notably, the representation dimension prior to the classification layer is set at 84, which will be utilized for extracting the representative subspace to compute the complementarity. In the FL training phase, we execute 50 communication rounds. Each round consists of local training iterations that vary depending on the dataset: 200 iterations for CIFAR-10 and 400 iterations for CIFAR-100.  Training employs the SGD optimizer with an initial learning rate 0.01 and a batch size 64. \\

\noindent \textbf{Hyperparameter Setup.} We have set key hyperparameters for optimal performance. We use three eigenvectors ($k = 3$) for our representative subspace. The regularization hyperparameter $\lambda$ is set at 1. Additionally, the hyperparameters $\alpha$ and $\beta$ control the degree of complementarity and similarity in our optimization equation. In experiments, we consider two scenarios based on client dataset characteristics. For datasets with complementarity, $\alpha=0.9$ and $\beta=1.4$ balance similarity and complementarity for enhanced performance. In contrast, for datasets lacking complementarity, such as in the Pathological partition, we reduce complementarity by setting $\alpha=0.5$ and $\beta=1.6$. The first setting is generally applied unless low complementarity among clients is known, in which case the second setting is used. To facilitate convergence, we use the initial settings for the first 70\% of communication rounds, then set $\alpha=0$ in the remaining rounds.


\subsection{Multimodal Implementation Setup} In our setup, we distribute the \emph{CUB200-2011} dataset among clients, with an equal split between image and text modalities. Each client possesses distinct feature extraction networks and uniform classification networks to fulfill the classification task. Within the same modality, we employ the unimodal \texttt{FedSaC} method to facilitate cooperation among clients. For cross-modality cooperation, structural differences in feature extraction layers necessitate restricting collaborative efforts to the classification layer. Given the inherent complementarity between different modalities, we focus on the similarity within the classification layers during cooperation. The cooperation weights for the classification layers are derived by excluding the complementarity term $\mathcal{C}$ from the optimization~\ref{eq:server}. These weights are used to aggregate the classification layers at the server side, enabling the effective fusion of cross-modal information.

\subsection{Multimodal Implementation Details} 
We allocate the CUB200-2011 dataset across 8 clients, with 4 handling image data and the others processing text data. For image modality clients, we employ a CNN architecture with four convolutional layers and a single classification layer. Text clients, on the other hand, utilize a TextCNN network consisting of five convolutional layers and a classification layer. The representation dimension is set at 256. The training involves 30 communication rounds, each comprising 200 iterations. We employ the Adam optimizer with an initial learning rate of 0.001. Throughout the training, we adopt a balanced setting for similarity and complementarity, with $\alpha=0.7$ and $\beta=1.2$, keeping the rest of the setup consistent with the unimodal \texttt{FedSaC}. 


\subsection{DataSets}
In our experiments, CIFAR-10, CIFAR-100~\cite{krizhevsky2009learning} and CUB200-2011~\cite{welinder2010caltech} are all public dataset. 

\noindent \textbf{CIFAR-10 and CIFAR-100.} The CIFAR-10 and CIFAR-100 datasets are key benchmarks in machine learning, each containing 60,000 32x32 color images. CIFAR-10 is categorized into 10 classes with 6,000 images per class, suitable for basic image recognition. CIFAR-100, offering a finer classification challenge, divides the same number of images across 100 classes, with 600 images per class. Both datasets, split into 50,000 training and 10,000 test images, are extensively used for evaluating image classification algorithms.\\

\noindent \textbf{CUB200-2011.} The CUB200-2011 dataset is specifically tailored for fine-grained visual categorization tasks, focusing on bird species identification. It consists of 11,788 images of 200 bird species, with both training and testing sets. Each species comes with a set of images that offer varying poses and backgrounds, providing a comprehensive dataset for advanced image recognition tasks. CUB200-2011 is particularly useful for research in areas requiring detailed visual discrimination, such as in distinguishing between closely related species.\\

\begin{figure*}[tbp]
\centering{
\includegraphics[width=\columnwidth]{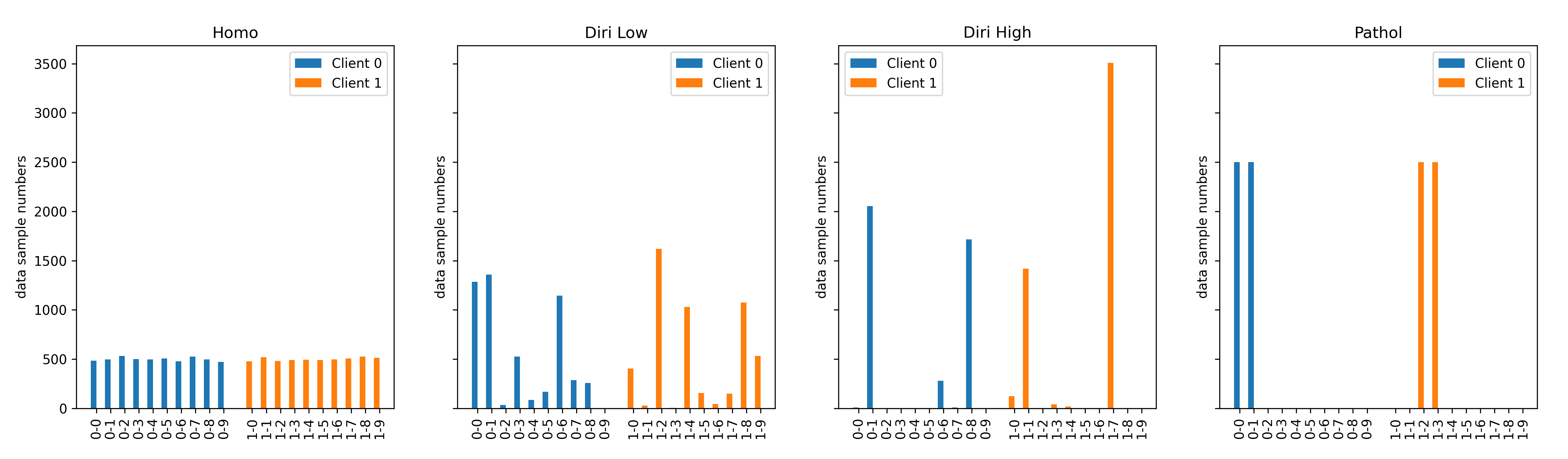}}
\caption{Illustration of the level of heterogeneity under four distinct partitioning schemes.}
\label{fig:supple_partition}
\end{figure*}

\noindent \textbf{Heterogeneity Partition.} In our study, we employ the CIFAR-10 dataset and select two clients to illustrate the level of heterogeneity under four distinct partitioning schemes, shown as~\ref{fig:supple_partition}.

\subsection{Baselines}
\noindent \textbf{FedAvg}~\cite{DBLP:conf/aistats/McMahanMRHA17} streamlines the training of deep networks from decentralized data in federated learning. It enables multiple clients to collaboratively train a shared model while maintaining data privacy and reducing communication overhead. Suitable for scenarios where central data collection is impractical due to privacy concerns, like in IoT and healthcare applications.\\
\vspace{-.3cm}

\noindent \textbf{FedProx}~\cite{DBLP:conf/mlsys/LiSZSTS20} specifically tackles system and statistical heterogeneity in federated networks. It introduces a proximal term to the optimization objective, enhancing stability and accuracy in networks with devices of varying capabilities. This modification leads to more robust convergence and improved accuracy in heterogeneous settings.\\
\vspace{-.3cm}

\noindent \textbf{CFL}~\cite{DBLP:journals/tnn/SattlerMS21}  is designed for large-scale peer-to-peer networks, optimizing federated learning by aggregating local model updates in a hierarchical manner. It ensures communication efficiency and data privacy through secure and authenticated encryption techniques. CFL stands out for its significant improvement in communication and computational efficiency, while robustly maintaining data integrity and privacy.\\
\vspace{-.3cm}

\noindent \textbf{pFedMe}~\cite{DBLP:conf/nips/DinhTN20}  introduces a personalized federated learning algorithm using Moreau envelopes as clients' regularized loss functions, allowing for the decoupling of personalized model optimization from global model learning. pFedMe is effective in handling statistical diversity among clients, leading to state-of-the-art convergence rates and superior empirical performance compared to traditional FedAvg and Per-FedAvg algorithms.\\
\vspace{-.3cm}

\noindent \textbf{Ditto}~\cite{DBLP:conf/icml/00050BS21} is a framework that enhances federated learning by simultaneously achieving fairness and robustness through personalization. It addresses the challenges of statistical heterogeneity in networks, using a simple yet scalable technique that improves accuracy, fairness, and robustness. Ditto is particularly effective against training-time data and model poisoning attacks and reduces performance disparities across devices.\\
\vspace{-.3cm}

\noindent \textbf{FedAMP}~\cite{DBLP:conf/aaai/HuangCZWLPZ21} This method employs federated attentive message passing to facilitate collaborations among clients with similar non-iid data, establishing convergence for both convex and non-convex models. FedAMP emphasizes pairwise collaborations between clients with similar data, overcoming the bottleneck of one global model trying to fit all clients in personalized cross-silo federated learning scenarios.\\
\vspace{-.3cm}

\noindent \textbf{FedRep}~\cite{DBLP:conf/icml/CollinsHMS21} utilizes a shared data representation across clients while allowing unique local heads for each client. This approach harnesses local updates concerning low-dimensional parameters, enabling efficient learning in heterogeneous data environments. By focusing on linear convergence and sample complexity, FedRep demonstrates improved performance over alternative personalized federated learning methods, especially in federated settings with non-iid data.\\
\vspace{-.3cm}

\noindent \textbf{pFedHN}~\cite{DBLP:conf/icml/ShamsianNFC21} introduces a personalized federated learning approach using hypernetworks. This method trains a central hypernetwork to generate unique personal models for each client, effectively sharing parameters across clients. It excels in handling data disparities among clients, reducing communication costs, and generalizing better to new clients with varying distributions and computational resources.\\
\vspace{-.3cm}

\noindent \textbf{FedRoD}~\cite{DBLP:conf/iclr/ChenC22}  simultaneously addresses generic and personalized learning objectives. It employs a two-loss, two-predictor system, decoupling the tasks of generic model training and personalized adaptation. The framework uses a class-balanced loss for the generic predictor and an empirical risk-based approach for the personalized predictor, facilitating robustness to non-identical class distributions and enabling zero-shot adaptation and effective fine-tuning for new clients.\\
\vspace{-.3cm}

\noindent \textbf{kNN-Per}~\cite{DBLP:journals/corr/abs-2111-09360} introduces local memorization using k-nearest neighbors in federated learning, enhancing the model's ability to personalize based on individual device data. This method stands out in its use of local data patterns to inform the federated learning process.\\
\vspace{-.3cm}

\noindent \textbf{pFedGraph}~\cite{DBLP:conf/icml/YeNWCW23} proposes the construction of inferred collaboration graphs among clients in federated learning. It dynamically computes these graphs based on the volume of data and model similarity at each client. This method strategically identifies similar clients for cooperation, effectively mitigating issues arising from data heterogeneity. \\
\vspace{-.3cm}

\noindent \textbf{FedIoT}~\cite{DBLP:conf/iotdi/ZhaoBH22} proposes a multimodal federated learning framework for IoT data, utilizing autoencoders to process multimodal data from clients. It introduces a multimodal FedAvg algorithm to aggregate local models from diverse data sources, enhancing classification performance in semi-supervised scenarios with unimodal and multimodal clients.

\subsection{Computing Resources}
Part of the experiments is conducted on a local server with Ubuntu 16.04 system.
It has two physical CPU chips which are Intel(R) Xeon(R) Gold 6248 CPU @ 2.50GHz with 20 cpu cores. The other experiments are conducted on a remote server. It has 8 GPUs which are GeForce RTX 3090.

\section{Privacy Discussion}
Our \texttt{FedSaC} exhibits similar data privacy preservation compared with baselines, as it does not share any private data of the clients. During communication, only model parameters are allowed to be shared. Similar to baselines, the sharing of model parameters is intended to maintain data privacy. The representative subspaces are derived from local data feature statistics generated by the model, a method that does not reveal any privacy details of the original dataset. Our approach is also compatible with protective strategies like differential privacy~\cite{DBLP:journals/tifs/RajkumarGLG22}. Specifically, for representative subspaces, we primarily rely on calculating their principal angles. Therefore, we could apply methods such as random cropping and adding minor noise to ensure that the original data cannot be reconstructed.

\section{Supplementary Experiments}
\label{Supplementary Experiments}
\subsection{Hyperparameters Experiments}
\begin{figure*}[tbp]
\centering{
\includegraphics[width=\columnwidth]{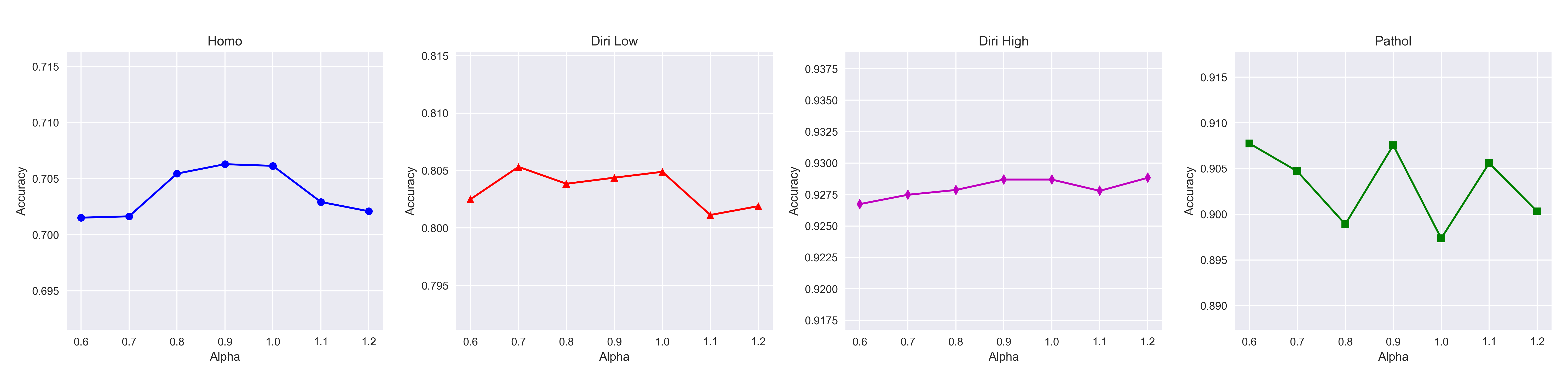}}
\caption{Average accuracy curves of the four partitions under various hyperparameter $\alpha$ settings. }
\label{fig:hyper_alpha}
\end{figure*}

\begin{figure*}[tbp]
\centering{
\includegraphics[width=\columnwidth]{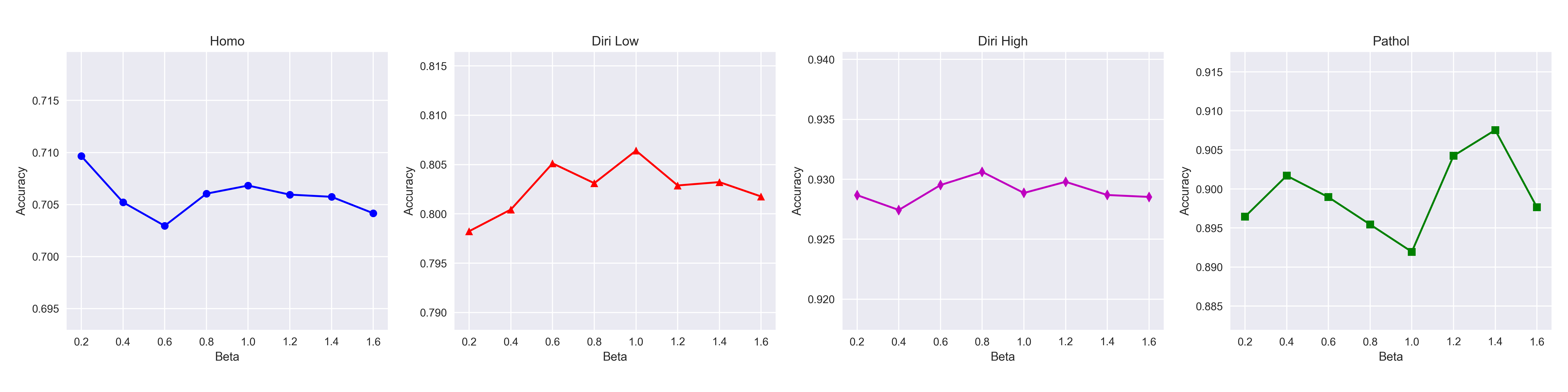}}
\caption{Average accuracy curves of the four partitions under various hyperparameter $\beta$ settings. }
\label{fig:hyper_beta}
\end{figure*}

The experimental analysis focused on evaluating the influence of hyperparameters, as illustrated in Figures~\ref{fig:hyper_alpha} and ~\ref{fig:hyper_beta}.

\noindent \textbf{Hyperparameter $\alpha$.}  In Figure~\ref{fig:hyper_alpha}, the similarity hyperparameter, denoted as $\beta$, is  fixed at 1.4, hile the complementary hyperparameter, $\alpha$, varied from 0.6 to 1.2. The results indicate that in data partitions characterized by complementarity, a moderate increase in $\alpha$ enhances accuracy. However, in partitions with high heterogeneity, the influence of $\alpha$  on the outcomes exhibits fluctuations. Notably, the experimental results consistently outperform the baseline, irrespective of the variations in $\alpha$. 

\noindent \textbf{Hyperparameter $\beta$.} Figure~\ref{fig:hyper_beta} presents the outcomes with $\alpha$ set at 0.9, examining the impact of changes in 
$\beta$ ranging from 0.2 to 1.6. It is observed that an optimal level of similarity substantially benefits the experimental results, which uniformly exceed the baseline performance. 

\noindent \textbf{Hyperparameter $\lambda$.} We employed the CIFAR100 dataset to assess the impact of the hyperparameter $\lambda$, associated with regularization constraints in local training, as demonstrated in Table~\ref{fig:hyper_lambda}. The results indicate that setting $\lambda$ to either 0.01 or 0.1 yields favorable outcomes with minimal fluctuation.
\begin{table*}[tbp]
\centering
\begin{tabular}{c|c c c c}
\hline
$\lambda$ & Homo & Diri(low) & Diri(high) & Pathol \\
\hline
0.005 &32.21 & 39.95 &55.79 &56.68 \\
0.01 &32.95 & 39.90 &55.91 &57.48 \\
0.05 &32.62 & 39.60 &55.81 &56.47 \\
0.1 &32.91 & 39.73 &56.04 &56.86 \\
0.2 &32.82 & 39.77 &55.53 &56.52 \\
\hline
\end{tabular}
\caption{Average accuracy of the four partitions under various hyperparameter $\lambda$ settings.}
\label{fig:hyper_lambda}
\end{table*}

\noindent \textbf{Hyperparameter $k$.} the influence of the subspace dimensionality, represented by $k$, on the experimental outcomes was examined, as detailed in Table~\ref{tab:k}. The findings suggest that $k=3$ is an appropriate choice for obtaining the representative subspace.
\begin{table*}[tbp]
\centering
\begin{tabular}{c|c c c c}
\hline
k & Homo & Diri(low) & Diri(high) & Pathol \\
\hline
1 &70.26 & 80.29 &92.97 &92.47 \\
2 &70.50 & 80.28 &93.04 &92.56 \\
3 &70.89 & 80.46 &93.14 &92.89 \\
4 &70.20 & 80.15 &92.73 &92.59 \\
\hline
\end{tabular}
\caption{Average accuracy of the four partitions under various hyperparameter $k$ settings.}
\label{tab:k}
\end{table*}

\subsection{Experiments in Large-Scale Client Cooperation}
\label{sec:large_scale}
In our experiments, we primarily focused on scenarios with a limited number of clients, specifically 8-10 clients. In situations involving a large number of clients, while the computational time overhead may not significantly impact our performance – a topic we will delve into in the following section – the cooperation among numerous clients could affect the convergence and stability of the collaboration. Therefore, for cooperation with a large client base, we incorporated an additional process to control the number of collaborators.

Specifically, in large-scale client cooperation, we randomly select k clients (k=10 in application) for collaboration before each iteration. This approach ensures convergence while enhancing cooperation efficiency. Table~\ref{tab:large-scale} presents the results in cooperation networks with 50 and 100 clients, incorporating this step. The results confirm the effectiveness of such collaborative efforts.

\begin{table}[ht]
\centering
\label{your-label}
\begin{tabular}{c|c|c c c c}
\hline
Modal                   & Client Num & Homo      & Diri(low)   & Diri(high)   & Pathol   \\ \hline
\multirow{2}{*}{Local}  & 50         & 8.63        & 22.45          & 45.36           & 33.33       \\
                        & 100        & 6.87        & 18.19          & 40.62           & 27.78       \\ \hline
\multirow{2}{*}{FedSaC} & 50         & 18.81        & 24.03         & 45.78           & 35.40       \\ 
                        & 100        & 11.90        & 19.11          & 40.75           & 28.51       \\ \hline
\end{tabular}
\caption{Large-scale client cooperation}
\label{tab:large-scale}
\end{table}

\vspace{-.3cm}
\section{Computational Cost and Complexity Analysis}
\label{sec:compute}
In comparison to classical federated learning methods, our approach incurs additional time overhead to provide informative feedback for client collaboration, aiming to obtain more suitable personalized models. The extra time expenditure primarily stems from three aspects: computing the similarity metric $\mathcal{S}$, computing the complementarity metric $\mathcal{C}$, and solving the optimization equation. We will analyze the computational complexity and demonstrate that this overhead is acceptable.

The computational complexity of the similarity measure $\mathcal{S}$ is proportional to the model parameters, approximating the cost of one inference time. For the complementarity measure $\mathcal{C}$, the feature matrix $X$ is first inferred. In cases of large local sample volumes, random sampling can approximate the local data distribution for effective computation of complementarity. Assuming random sampling of $m$ samples with feature dimension $d$, the resulting feature matrix $X \in \mathbb{R}^{m \times d}$ is processed. The complexity of extracting the representative subspace using SVD is $O(md^2)$. In practical computations, this step's overhead is almost negligible compared to the duration of model training.

The subsequent step involves solving the optimization equation. Notably, each row of the adjacency matrix $W$ in Equation~\ref{eq:server} is independent, allowing for the independent computation of cooperation weights for each client with others. We simplify the optimization equation as follows:

\begin{equation}
\begin{split}
\min\limits_{\boldsymbol{W}_{i*}}&\sum_{j=1}^N\Biggl(\boldsymbol{W}_{ij}^2 - 2p^j\boldsymbol{W}_{ij} + (p^j)^2 + \alpha\boldsymbol{W}_{ij}\cos\left(\frac{1}{k}\cdot\sum_l\phi_l\right) - \beta\boldsymbol{W}_{ij}\frac{\boldsymbol{\theta}^i\cdot\boldsymbol{\theta}^j}{\|\boldsymbol{\theta}^i\|\cdot\|\boldsymbol{\theta}^j\|}\Biggr)
\\ & \text{s.t.} \quad \quad \quad \quad \sum_{j=1}^N\boldsymbol{W}_{ij}=1,\forall{i};\quad \boldsymbol{W}_{ij}\geq 0,\forall{i, j} 
\end{split}
\vspace{-.3cm}
\end{equation}
It can be deduced that:

\begin{equation}
\begin{split}
\min\limits_{\boldsymbol{W}_{i*}}&\sum_{j=1}^N\Biggl(\boldsymbol{W}_{ij}^2 +\Biggl(\alpha\cos\left(\frac{1}{k}\cdot\sum_l\phi_l\right) - \beta\frac{\boldsymbol{\theta}^i\cdot\boldsymbol{\theta}^j}{\|\boldsymbol{\theta}^i\|\cdot\|\boldsymbol{\theta}^j\|} -  2p^j\Biggr)\boldsymbol{W}_{ij}\Biggr)
\\ & \text{s.t.} \quad \quad \quad \quad \sum_{j=1}^N\boldsymbol{W}_{ij}=1,\forall{i};\quad \boldsymbol{W}_{ij}\geq 0,\forall{i, j} 
\end{split}
\vspace{-.3cm}
\end{equation}

It is evident that the objective of the optimization equation is equivalent to $\sum_{j=1}^N(\boldsymbol{W}_{ij}^2 + \phi_{ij}\boldsymbol{W}_{ij})$, which forms a quadratic optimization function. This function is convex, as evidenced by its compliance with the convex set inequality constraint $\boldsymbol{W}_{ij}\geq 0, \forall{i, j}$ and the affine transformation equality constraint $\sum_{j=1}^N\boldsymbol{W}_{ij}=1, \forall{i}$. Therefore, this optimization problem is a convex optimization problem, solvable using convex optimization solvers. Such solvers can rapidly find the unique optimal solution.

We tested the runtime of each additional phase in our experiments on our platform and compared it with the training duration of a single client in one local training round, given ten clients, as shown in Table~\ref{tab:time}. The results indicate that the time required for similarity metric and solving the optimization equation is negligible compared to local training duration. Although the complementarity metric phase, which involves an inference process, does take some time, it is still significantly less than the local training duration. Therefore, the additional cost of cooperation is acceptable. As a result, \texttt{FedSaC} does not introduce substantial additional computational and communication costs, making its computational overhead comparable to existing baselines.
\begin{table*}[tbp]
\centering
\begin{tabular}{c|c}
\hline
Phases & Run-time Consumption  \\
\hline
Local Training & 21.21s\\
Similarity  Metric & 0.05s\\
Complementarity  Metric & 1.62s\\
Optimization Equation & 0.06s\\
\hline
\end{tabular}
\caption{Run-time consumption of across different phases.}
\label{tab:time}
\vspace{-.6cm}
\end{table*}

\vspace{-.3cm}
\section{Convergence Analysis}
\begin{figure*}[htbp]
\centering{
\includegraphics[width=0.6\columnwidth]{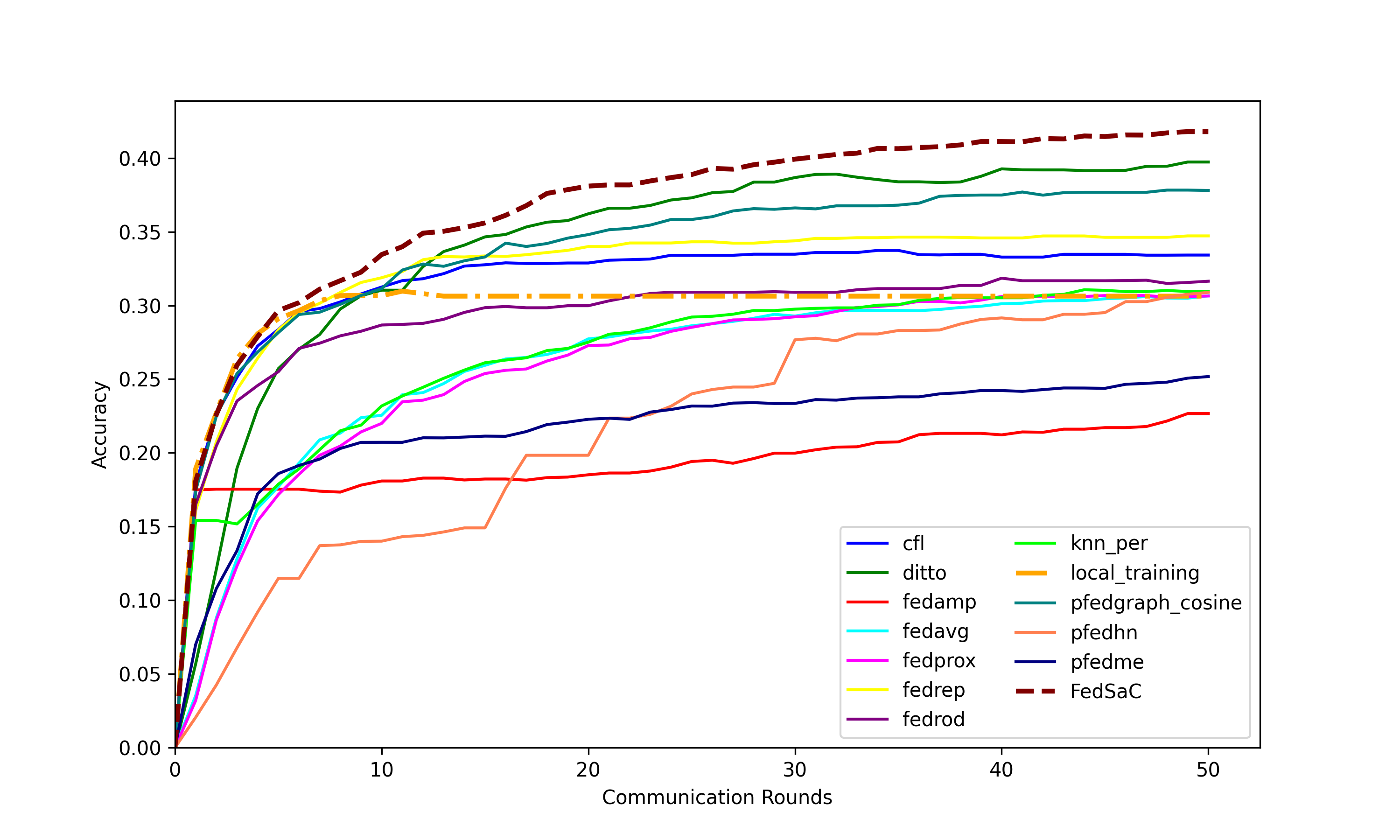}}
\caption{Illustration of the accuracy progression over communication rounds on the CIFAR-100 dataset under a Diri(low) partition.}
\label{fig:convergence}
\end{figure*}

The introduction of complementarity in our \texttt{FedSaC} approach does not lead to convergence issues. As depicted in Figure~\ref{fig:convergence}, we illustrate the accuracy progression over communication rounds on the CIFAR-100 dataset under a Diri(low) partition. It is observed that the accuracy of the FedSaC method steadily rises and gradually converges. Unlike local training, which may lead to overfitting and a subsequent decline in accuracy due to excessive training, our method effectively circumvents the overfitting problem. In contrast to other baselines that converge prematurely and potentially get trapped in local optima, our approach consistently explores better solutions, achieving optimal performance before ultimately converging.

\end{document}